\documentclass{article}
\usepackage{iclr2027_conference,times}

\usepackage{amsmath,amsfonts,bm}

\def\eqref#1{equation~\ref{#1}}

\def\1{\bm{1}}

\DeclareMathAlphabet{\mathsfit}{\encodingdefault}{\sfdefault}{m}{sl}
\SetMathAlphabet{\mathsfit}{bold}{\encodingdefault}{\sfdefault}{bx}{n}

\usepackage{amsmath}
\usepackage{amssymb}
\usepackage{booktabs}
\usepackage{float}
\usepackage{graphicx}
\usepackage{multirow}
\usepackage{array}
\usepackage[table]{xcolor}
\usepackage{tcolorbox}
\tcbuselibrary{skins,breakable}

\definecolor{caseinput}{HTML}{1F6FB2}
\definecolor{caseprobe}{HTML}{2E7D32}
\definecolor{casemixed}{HTML}{8E44AD}

\newtcolorbox[auto counter]{probecase}[2][]{
    enhanced,
    breakable,
    colback=white,
    colframe=black!45,
    colbacktitle=black!5,
    coltitle=black,
    title={Example~\thetcbcounter: #2},
    fonttitle=\bfseries\small,
    boxrule=0.5pt,
    titlerule=0pt,
    arc=1.2mm,
    left=2mm, right=2mm, top=2mm, bottom=2mm,
    toptitle=2mm, bottomtitle=2mm,
    before upper={\setlength{\parindent}{0pt}},
    before skip=10pt, after skip=10pt,
    #1
}

\newtcolorbox{probepanel}[3][]{
    enhanced,
    colback=white,
    colframe=#2!65!white,
    colbacktitle=#2!10!white,
    coltitle=#2!85!black,
    title={#3},
    fonttitle=\bfseries\small,
    fontupper=\small,
    fontlower=\small,
    boxrule=0.5pt,
    titlerule=0pt,
    arc=1.2mm,
    left=3mm, right=3mm, top=2mm, bottom=2mm,
    toptitle=2mm, bottomtitle=2mm,
    segmentation style={solid, draw=#2!30!white, line width=0.5pt},
    before upper={\setlength{\parindent}{0pt}},
    before lower={\setlength{\parindent}{0pt}},
    before skip=2pt, after skip=6pt,
    #1
}
\usepackage{microtype}
\usepackage{enumitem}
\usepackage{hyperref}
\usepackage{url}

\title{MoMaTune: Model, Match, Then Tune\\Language Models for Numerical Records}

\iclrfinalcopy 

\author{\mdseries
Weimin Lyu\textsuperscript{1} 
Jingfen Zhang\textsuperscript{1} 
Lingdao Sha\textsuperscript{1} 
Peng Sun\textsuperscript{1} 
Xiaowei Chen\textsuperscript{1}  
Vincent Gao\textsuperscript{1} \\
Rui Song\textsuperscript{1}
\\
\centering
\makebox[\textwidth][c]{\textsuperscript{1}Amazon}
}

\newcommand{\methodname}{\textsc{MoMaTune}}
\newcommand{\modelstage}{\textsc{Model}}
\newcommand{\matchstage}{\textsc{Match}}
\newcommand{\tunestage}{\textsc{Tune}}
\newcommand{\recipedefs}{\modelstage{} denotes next-token record adaptation; \matchstage{} denotes optional same-record consistency; and \tunestage{} denotes supervised fine-tuning (SFT).}
\newcommand{\kv}{\texttt{<key: value>}}
\newcommand{\recordloss}{\mathcal{L}_{\mathrm{record}}}
\newcommand{\consloss}{\mathcal{L}_{\mathrm{cons}}}

\newcommand{\best}[1]{\textbf{#1}}

\newcolumntype{Y}{>{\raggedright\arraybackslash}p{0.19\textwidth}}

\newcommand{\myparagraph}[1]{\smallskip\noindent\textbf{#1}}

\makeatletter
\newcommand{\inlinefigurecaption}[1]{%
  \refstepcounter{figure}%
  \vskip\abovecaptionskip
  {\raggedright\noindent\fnum@figure: #1\par}%
  \vskip\belowcaptionskip}
\makeatother

\let\originalsubsection\subsection
\RenewDocumentCommand{\subsection}{s o m}{%
  \IfBooleanTF{#1}{\originalsubsection*{#3}}{%
    \IfNoValueTF{#2}{\originalsubsection{#3}}{\originalsubsection[#2]{#3}}}%
  \def\currentsubsectiontitle{#3}%
  \def\computingcoststitle{Computing Costs}%
  \ifx\currentsubsectiontitle\computingcoststitle
    \global\catcode`\_=12\relax
    \global\let\originalparagraphend\par
    \gdef\par{%
      \global\catcode`\_=8\relax
      \global\let\par\originalparagraphend
      \originalparagraphend
    }%
  \fi
}

\begin{document}
\maketitle

\begin{abstract}
Despite remarkable progress in language understanding and generation, decoder language models continue to exhibit striking failures on numerical data.  
Numerical examples are typically serialized as text by concatenating feature names with their values (e.g., \texttt{fat:3.4}), while conventional task adaptation applies loss only to the downstream prediction target.
This leaves two structures weakly supervised: \emph{within-value sensitivity}, where digit position determines numerical magnitude, because token cross-entropy contains no explicit numerical metric; and \emph{across-field structure}, where measurements in the same record are statistically dependent, because target-only tuning provides no direct loss on the measurements or their dependencies.
We introduce \methodname{} (\textbf{Mo}del, \textbf{Ma}tch, then \textbf{Tune}), a two-stage post-training strategy organized around these supervision gaps. Stage 1 applies \modelstage{}, ordinary next-token prediction densely over label-free
serialized records, and optionally adds \matchstage{}, a same-record consistency objective that aligns representations of different partial views of the same example. Stage 2 applies \tunestage{}, supervised fine-tuning for downstream target prediction.
We evaluate \methodname{} on 169,285 records from eight public datasets, with 339 of 340 input fields numerical. At 1.5B parameters, \modelstage{} raises the eight-dataset masked-value imputation aggregate from 0.106 to 0.401, while the three-dataset correct-context gain rises from 0.058 to 0.167. On diagnostics, \modelstage{} also increases sensitivity to numerically consequential digit changes, while \matchstage{} improves robustness to informative partial views. Adding \tunestage{} yields higher target-supervised embedding aggregates for each tested Stage-1 parent, and the two-stage recipes obtain higher aggregate scores than \tunestage{} alone. The same pattern is observed at 4B, 9.6B, and 27B parameters when model embeddings are used for downstream prediction.

\end{abstract}

\section{Introduction}
\label{sec:introduction}
Many scientific and industrial examples contain little or no natural language; they consist primarily of numerical measurements. To process them with decoder-only LMs such as Qwen~\citep{bai2023qwen}, a common approach is to serialize named measurements as attribute-value pairs, for example \texttt{<protein:6.2> <fat:3.4>}. We call each pair a \emph{field} and the resulting sequence a \emph{concatenated record}. Although this serialization makes numerical examples compatible with pretrained LMs, the numerical meaning of each value and the relationships among measurements remain implicit in the token sequence.

Prior work uses such serializations with task-specific supervision for prediction and  generation~\citep{tabllm2023,great2023,gtl2024,tabula2024,tabgemma2025}. Complementary methods modify tokenization or training objectives to encode magnitude, order, or feature structure more explicitly~\citep{xval2023,tpberta2024,subtab2021,rankncontrast2023}. These directions demonstrate that LMs can operate on structured data, but leave two questions unresolved: \textit{Can a standard causal decoder learn numerical significance within values and conditional relationships across fields?}

\begin{figure}[t]
    \vspace{-.2in}
    \centering
    \includegraphics[width=\textwidth]{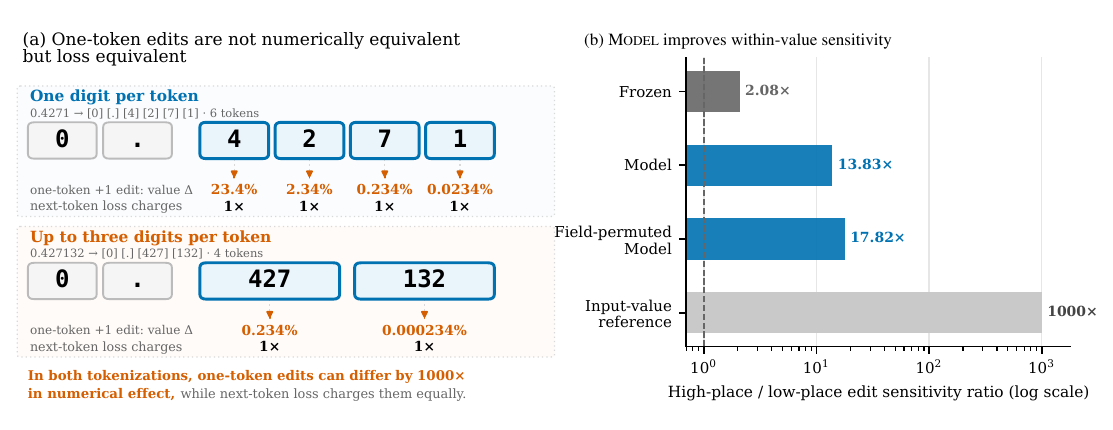}
    \vspace{-.3in}
    \caption{\textbf{Token loss is numerically agnostic, but record-level adaptation improves magnitude sensitivity than target-only SFT.}
    (a) With one digit per token, incrementing the first versus fourth fractional digit of
    \texttt{0.4271} produces a $1000\times$ difference in numerical effect. With groups of up to three digits, incrementing \texttt{[427]} versus \texttt{[132]} by one in \texttt{0.427132} also produces a $1000\times$ difference. Each intervention changes exactly one token. 
    (b) We measure within-value sensitivity as the ratio between how far the pooled record embedding moves after a first-fractional-digit edit and after a fourth-fractional-digit edit. On the MiniBooNE diagnostic, applying the same objective densely through field-permuted \modelstage{} increases this ratio from 2.08$\times$ to 17.82$\times$ at 1.5B. The $1000\times$ difference in numerical effect is a reference, not a required embedding-distance ratio.
    \modelstage{} denotes next-token prediction on records, and field-permuted means reordering the attribute-value fields as data augmentation.}
    \label{fig:place-value}
    \vspace{-.2in}
\end{figure}

In this paper, we identify two structural requirements of numerical records that help explain current LM numeracy weaknesses and examine them with complementary diagnostics. First, \emph{within-value sensitivity} requires distinguishing digit positions by their effect on numerical magnitude. Fig.~\ref{fig:place-value}(a) shows why digit position matters: in \texttt{0.4271}, increasing the first digit after the decimal changes the value by 23.4\%, whereas increasing the fourth changes it by only 0.0234\%, a $1000\times$ difference. Appx.~\ref{app:digit-blindness} reports the tokenizer survey. Token cross-entropy is numerically agnostic: it provides no explicit weighting by numerical consequence, although dense record-level prediction can still learn such structure from the data distribution. In a MiniBooNE digit-perturbation diagnostic, the frozen 1.5B model is only $2.08\times$ more sensitive to the first-digit edit than to the fourth-digit edit (Fig.~\ref{fig:place-value}(b)). Second, \emph{across-field structure} requires using relationships among measurements in the same record. If $\mathbf{x}=(x_1,\ldots,x_d)$, predicting a missing field $x_j$ requires modeling $p(x_j\mid\mathbf{x}_{-j})$, rather than only the marginal distribution of $x_j$. Across the eight datasets, the frozen 1.5B model obtains a masked-value imputation aggregate of 0.106, below the 0.212 field-conditional median baseline (Tab.~\ref{tab:imputation-completion-1p5b}). In a three-dataset context diagnostic, its correct-versus-mismatched context gain is only 0.058
(Tab.~\ref{tab:cross-field-context}). Together, these complementary diagnostics show limited sensitivity to numerical magnitude in the digit-position test and limited use of cross-field context in masked-value completion.

These two challenges motivate three separable post-training roles: learn numerical and conditional
record structure, optionally preserve useful information across partial views when inputs may be
incomplete or reordered, and expose the resulting representation for target prediction. The
second role is deployment-dependent rather than universally beneficial.

We therefore propose \methodname{}, a modular two-stage post-training framework whose core path is \modelstage{}$\rightarrow$\tunestage{}. Stage 1 applies \modelstage{}, ordinary next-token prediction densely over every visible token of label-free numerical records, to strengthen within-value sensitivity, record completion, and use of across-field context. For settings that require robustness to partial records, Stage 1 can additionally include \matchstage{} (same-record consistency), a regularizer that brings representations of alternative partial views of the same example closer together. We study both disjoint field subsets and \emph{permuted-prefix views}, formed by independently permuting field order. Stage 2 applies \tunestage{} (SFT): answer-masked supervised fine-tuning in a fixed field order for downstream target prediction. Thus, \modelstage{}$\rightarrow$\tunestage{} forms the core training path, while \matchstage{} is an optional robustness component whose effect depends on the chosen partial-view construction.
One robustness-oriented variant \methodname{}, uses field-permuted \modelstage{} with \matchstage{}, followed by fixed-order \tunestage{}. \matchstage{} draws on unified generative--representational training \citep{grit2024}, same-instance subset learning \citep{subtab2021}, and representation consistency, but uses neither labels nor negative mining.

We evaluate \methodname{} through frozen-embedding prediction and direct decoder tests across eight public datasets containing 169,285 records. The evaluated dataset is almost entirely numerical: 339 of 340 input fields are numerical, and seven of the eight datasets contain
only numerical input fields. Our contributions are threefold:

\begin{itemize}
    \item \textbf{Identify two numerical-record challenges.} We identify and empirically diagnose
    \emph{within-value sensitivity} and \emph{across-field structure} as two requirements that
    current decoder LMs weakly capture: models must represent how digit position determines
    magnitude and how measurements within the same record depend on one another.
    \item \textbf{Introduce a modular post-training strategy.} The core \modelstage{}$\rightarrow$\tunestage{} path first adapts the decoder to numerical-record structure and then specializes the resulting representation for target prediction. \matchstage{} is an optional same-record consistency regularizer for partial-view robustness whose utility depends on the paired views.
    \item \textbf{Evaluate numerical-record competence comprehensively.} We evaluate embedding
    and generation tasks on a near-purely numerical benchmark across multiple
    datasets and model scales, using matched controls to isolate the contribution of each training
    stage and data-view choice.
\end{itemize}


\section{Related Work}
\label{sec:related}

\myparagraph{Numerical representation learning.}
Standard LM objectives do not explicitly encode numerical distance or place value, and numerical representations can remain weak on magnitude and arithmetic tests \citep{wallace2019,numeracygap2025}. Prior work adds numerical structure through modified tokenization, feature-aware pretraining, subset views, or target-distance objectives \citep{xval2023,tpberta2024,subtab2021,rankncontrast2023}. We instead ask whether a standard causal decoder, with unchanged tokenization and attention, can learn numerical magnitude and cross-field structure directly from label-free serialized records.

\myparagraph{Joint generation and representation learning.}
GRIT jointly trains generation and representations \citep{grit2024}, while E5-Mistral, LLM2Vec, and NV-Embed develop decoder-derived encoders using supervised pairs, modified attention, specialized pooling, or embedding-focused training \citep{e5mistral2024,llm2vec2024,nvembed2024}. Our goal is instead to improve numerical representations while preserving causal decoding, then specialize them for downstream prediction.

\myparagraph{Serialized-record language modeling.}
TabLLM, GReaT, GTL, and TabuLa apply language models to serialized records for prediction or generation \citep{tabllm2023,great2023,gtl2024,tabula2024}. These works establish the usefulness of serialized records, but do not directly study what numerical structure a causal decoder can acquire from label-free record modeling before target supervision. Our work focuses on this question and evaluates how \modelstage{} strengthens numerical-record representations before \tunestage{} specializes them.

\section{\methodname: A Post-Training Framework for Numerical Records}
\label{sec:method}

    \subsection{Definition: Records, serialization, and embeddings}

    A numerical record $r$ is one dataset example containing $m$ fields:
    $r=((k_1,v_1),\ldots,(k_m,v_m))$. Each field stores one measurement as a key--value pair:
    $k_j$ names the measured variable, and $v_j$ is its observed numerical value in this record.
    Serialization maps this ordered field sequence to text. Preserving the dataset's column order,
    we define the serialized record as
    \begin{equation}
        x(r) = \texttt{<}k_1\texttt{: }v_1\texttt{>}\ \cdots\
               \texttt{<}k_m\texttt{: }v_m\texttt{>}.
    \end{equation}
    The downstream target $y$ is excluded from Stage 1 and introduced only in Stage 2. For embedding evaluation, we represent the serialized record by the masked mean of final-layer hidden states,
    \begin{equation}
        z(x)=\frac{\sum_{t=1}^{T}a_t h_t}{\sum_{t=1}^{T}a_t},
    \end{equation}
    where $a_t$ is the attention mask. Mean pooling is fixed across trained and frozen decoder arms.

\begin{figure}[t]
\vspace{-.2in}
    \centering
    \includegraphics[width=\textwidth]{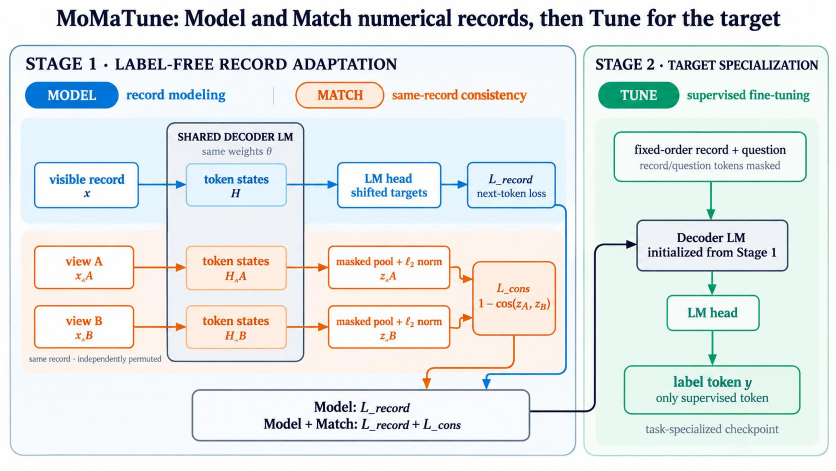}
    \vspace{-.4in}
    \caption{\textbf{\methodname{} centers on \modelstage{}$\rightarrow$\tunestage{}, with \matchstage{} as an optional partial-view robustness component.} Stage 1 always applies \modelstage{} to label-free numerical records. Applications that require robustness to partial records can additionally apply \matchstage{}. The resulting Stage-1 checkpoint initializes Stage 2, where \tunestage{} presents a fixed-order record and target question and applies loss only to the label token.}
    \label{fig:method}
    \vspace{-.15in}
\end{figure}

\noindent
\begin{minipage}[t]{.62\textwidth}
    \vspace{-.2in}
    \centering
    \includegraphics[width=\linewidth]{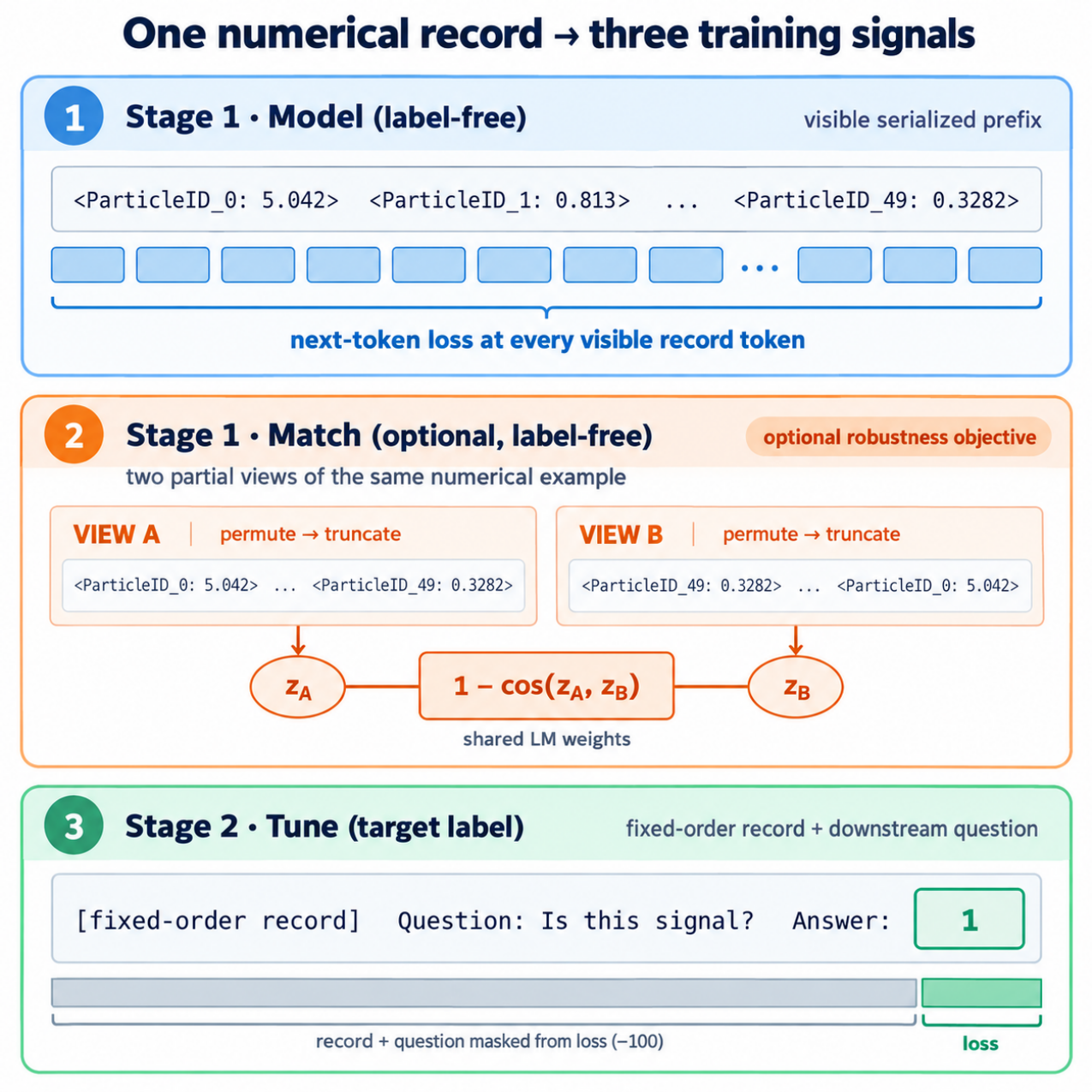}
    \vspace{-.2in}
    \inlinefigurecaption{\textbf{One numerical record supports three available training signals.} \modelstage{} applies
    next-token loss to every visible token in the truncated, label-free record. When partial-view
    robustness is desired, \matchstage{} encodes two label-free permuted-prefix views
    with the same decoder and minimizes their cosine distance, without labels or negative pairs.
    \tunestage{} returns to one fixed field order, masks the record and question tokens, and
    supervises only the label token.}
    \label{fig:examples}
    \vspace{-.15in}
\end{minipage}\hfill
\begin{minipage}[t]{.35\textwidth}
    \vspace{-.3in}

    Together, $x(r)$, $z(x)$, and $y$ provide the common notation for the three training signals shown
    in Fig.~\ref{fig:examples}. \modelstage{} applies next-token loss to the visible tokens of $x(r)$;
    \matchstage{} aligns the representations of two label-free views of the same record without
    observing $y$; and \tunestage{} appends a target question to $x(r)$, and suvervises on $y$. Fig.~\ref{fig:method} shows
    how these signals form the two-stage recipe; the following subsections define each stage and its
    objective.


\refstepcounter{subsection}
\subsection*{\thesubsection\hspace{0.4em}Stage 1: Record adaptation}
\addcontentsline{toc}{subsection}{\protect\numberline{\thesubsection}Stage 1: Record adaptation}

\paragraph{\modelstage{}: next-token record adaptation.}
\modelstage{} keeps ordinary causal next-token prediction but changes where it is applied: before target supervision, every visible token in a label-free record contributes to the loss:
\begin{equation}
    \hspace{-1em}\recordloss(x)=-\sum_{t=1}^{T}\log p_\theta(x_t\mid x_{<t}).
\end{equation}
This does not change the loss geometry; it supplies dense in-domain supervision for numerical values and their field context while preserving causal generation.

\end{minipage}
\par\vspace{.5\baselineskip}

\paragraph{\matchstage{}: conditional same-record consistency.}
We evaluate two constructions of positive views. In the fixed-order arms, the keys are randomly
partitioned into a larger first view (approximately 50--75\% of fields) and a disjoint second view;
fields within each view remain in their original order. In the field-permuted arms, the field order
is independently permuted twice and each serialization is then truncated at 512 tokens. We call
these \emph{permuted-prefix views}. They may overlap, but field-order permutation before truncation moves
different key-value units into the visible prefix. Thus paired views generally differ in both
order and visible content rather than preserving every value; Appx.~\ref{app:training-details}
reports the direct audit. After mean pooling and $\ell_2$ normalization, both minimize
\begin{equation}
    \consloss(r)=1-\hat z(x(r_A))^\top \hat z(x(r_B)).
\end{equation}
This asks two label-free views of the same example to have similar embeddings. There are no in-batch
negatives, or variance regularizers. \modelstage{} alone minimizes $\recordloss$. When partial-view robustness is desired, the model-and-match arm
instead uses
\begin{equation}
    \mathcal{L}_{\mathrm{Model+Match}}=\lambda_1\,\recordloss + \lambda_2\,\consloss.
    \label{eq:adapt}
\end{equation}
The choice is therefore between a core \modelstage{} checkpoint and a
\modelstage{}+\matchstage{} checkpoint, rather than a requirement that every Stage-1 run use both
losses. The consistency-only control drops $\recordloss$ and directly tests whether \modelstage{}
prevents collapse. A \modelstage{}+\matchstage{} step uses three backbone forwards (visible record,
view A, and view B),
versus one for the \modelstage{} baseline;
therefore a null result cannot be attributed to giving the auxiliary arm less computation.

\paragraph{Field-order permutation: data augmentation.}
By default, each row is serialized once in the dataset's original column order. This fixed
presentation places every field in the same position and local context, limiting data diversity and
encouraging positional shortcuts. Field-order permutation addresses this by reordering complete
key--value fields while preserving their names and values; multiple field-order permutations present
the same measurements in varied contexts and encourage learning field identities and cross-field
relationships without relying on a particular position.

\subsection{Stage 2: Target specialization}

\paragraph{\tunestage{}: supervised fine-tuning.}
Stage 1 is label-free and does not optimize for a downstream target. \tunestage{} therefore
specializes the Stage-1 checkpoint with SFT: each example concatenates $x(r)$, a target question,
and answers $y$; the record and question positions are masked, so only the answer tokens contribute to the loss. By default, \tunestage{} uses one canonical field order and
one view per record; field-permuted controls is not applied in this stage. Because \tunestage{} observes $y$, its embedding score measures supervised target specialization rather than label-free competence.

\section{Experimental Design}
\label{sec:experiments}

\subsection{Experimental Settings}

\myparagraph{Datasets.} We evaluate eight public datasets. Seven datasets are downloaded through OpenML
\citep{openml2014} and contain no text fields. They account for 161,389 of 169,285 records (95.3\%).
USDA FoodData is the only mixed-modality dataset: it adds one natural-language description to 48
numeric nutrients. Across the eight schemas, 339 of 340 input fields are numerical. All datasets use
fixed 80/10/10 train/validation/test splits with seed 42. Appx.~\ref{app:repro} summarizes the
corpus in Tab.~\ref{tab:datasets} and gives the complete construction and split statistics.

\myparagraph{Models.} Our primary experiments use Qwen2.5-1.5B-Instruct \citep{bai2023qwen}. To test
whether the findings persist at larger scales, we additionally evaluate Qwen3-4B-Instruct-2507,
Qwen3.5-9B (9.6B parameters), and Qwen3.8-27B \citep{qwen3}. Because these checkpoints differ in
architecture and tokenizer vocabulary, we treat this comparison as a model-size robustness sweep
across checkpoint families. Appx.~\ref{app:model-architectures} summarizes the model
architectures and tokenizers; Appx.~\ref{app:training-details} and~\ref{app:arms} provide the
optimization settings and complete arm registry.

\myparagraph{Evaluation.} Our primary evaluation measures whether post-training makes record embeddings useful for downstream target prediction. At evaluation, for each model checkpoint, we get the frozen vector per record and fed to standardized linear probe: logistic regression evaluated by ROC--AUC on two classification datasets and Ridge regression evaluated by $R^2$ on six regression datasets. Decoder LMs and trained recipes use mean-pooled final-layer hidden states; embedding-specialized models use their native pooling.
As complementary direct-decoder evidence, masked-value imputation
asks the model to recover a held-out field from the visible record; its primary metric is within-10\% accuracy, summarized across datasets by \textsc{Gen-8}.
Appx.~\ref{app:metrics} defines every metric, protocol, and dataset-to-metric mapping.


\subsection{Experiment Results}
\label{sec:results}

Sec.~\ref{sec:overall-target-performance} validates that our two-stage training framework obtains higher aggregate target-prediction scores than both the frozen backbone and \tunestage{} alone.
sec.~\ref{sec:distinct-capabilities} then separates the stages' roles: \modelstage{} improves numerical sensitivity and record completion,
\matchstage{} improves incomplete-record robustness only when paired views preserve compatible evidence;
and \tunestage{} specializes the representation for target prediction.
Sec.~\ref{sec:ablations} tests whether the our two-stage framework gains persist from 1.5B to 27B, and other ablations.


\subsubsection{Overall performance: target prediction and numerical completion}
\label{sec:overall-target-performance}


\myparagraph{Target prediction.}
Tab.~\ref{tab:embedding-per-dataset} shows that applying \modelstage{} before \tunestage{} provides the central target-prediction increase. In the fixed-order setting, \textsc{Emb-8} rises from 0.688 for the frozen checkpoint to 0.736 after \modelstage{} $\rightarrow$ \tunestage{}.\footnote{The 1.5B recipes underperform the 7B frozen decoder on HIGGS and Superconductivity, but our 9.6B \modelstage{} checkpoint obtains higher scores on both (Tab.~\ref{tab:model-size-embedding-full}).}
Adding \matchstage{} to the fixed-order pipeline yields a higher \textsc{Emb-8} of 0.743. Field-order permutation further raises the two-stage scores, from 0.736 to 0.753 without \matchstage{} and from 0.743 to 0.753 with \matchstage{} (Tab.~\ref{tab:embedding-per-dataset}; Appx. Tab.~\ref{tab:m24-m26}). 
Thus, the field-permuted \modelstage{} $\rightarrow$ \tunestage{} and field-permuted \modelstage{} $+$ \matchstage{} $\rightarrow$ \tunestage{} variants obtain the same aggregate \textsc{Emb-8} of 0.753, showing that \matchstage{} is not required for the highest complete-record aggregate in this comparison. 
Within the field-permuted pair, the variant without \matchstage{} obtains higher scores on five of eight individual datasets (Appx. Tabs.~\ref{tab:aggregate-ablation-controls} and~\ref{tab:m24-m26}). Together, these comparisons associate the main target-prediction increase with applying two-stage training, with field-order permutation providing an additional increase in the reported aggregate. We evaluate \matchstage{} separately below for its conditional partial-view robustness effect.

\begin{table*}[t]
\vspace{-.2in}
\caption{Our \methodname{} improves linear-probe target prediction
across eight nearly pure-numerical datasets.
Classification tasks report ROC--AUC, the regression tasks report $R^2$. \textsc{Emb-8} is unweighted mean of all eight dataset.
The two field-permuted variants obtain the highest reported \textsc{Emb-8} (0.753), with and without \matchstage{}.
All trained recipes in this table use a Qwen2.5-1.5B backbone;
Tab.~\ref{tab:model-size-scaling-primary} reports the corresponding scale comparison from 1.5B
to 27B. Additional classification metrics appear in Appx.
Tab.~\ref{tab:embedding-detail-cls}, and regression metrics in
Tabs.~\ref{tab:embedding-detail-r2}, \ref{tab:embedding-detail-mae},
and~\ref{tab:embedding-detail-pearson}.}
\label{tab:embedding-per-dataset}
\centering
\scriptsize
\setlength{\tabcolsep}{3pt}
\resizebox{\textwidth}{!}{%
\begin{tabular}{lrrrrrrrrr}
\toprule
\multirow{2}{*}{Model / recipe}
& \multicolumn{2}{c}{Classification: ROC--AUC $\uparrow$}
& \multicolumn{6}{c}{Regression: $R^2 \uparrow$}
& \multirow{2}{*}{Emb-8 $\uparrow$} \\
\cmidrule(lr){2-3}\cmidrule(lr){4-9}
& MiniBooNE & HIGGS & Supercond. & YearPred & Naval & CPU act.
& Protein & USDA & \\
\midrule
\multicolumn{10}{l}{\textit{Frozen decoder LMs (mean-pooled hidden states)}} \\
Qwen3-4B & .960 & .642 & .783 & .160 & .893 & .926 & .332 & .940 & .705 \\
Qwen3-8B & .955 & .636 & .797 & .172 & .917 & .931 & .330 & .946 & .711 \\
Qwen3.5-9B & .954 & .644 & .807 & .149 & .897 & .928 & .349 & .939 & .708 \\
Qwen3-14B & .959 & .660 & .809 & .156 & .929 & .947 & .342 & .951 & .719 \\
OLMo-2-7B & .961 & \textbf{.725} & \textbf{.817} & .160 & .930 & .921 & .329 & .941 & .723 \\
DeepSeek-V2-Lite (16B MoE) & .958 & .646 & .778 & .167 & .837 & .921 & .334 & .926 & .696 \\
gpt-oss-20b & .954 & .668 & .753 & .153 & .830 & .803 & .284 & .915 & .670 \\
Gemma-4-12B & .957 & .639 & .813 & .144 & .921 & .922 & .353 & .952 & .713 \\
Llama-3.1-8B & .960 & .712 & .804 & .174 & .915 & .911 & .321 & .946 & .718 \\
\midrule
\multicolumn{10}{l}{\textit{Embedding-specialized models (model-specific pooling)}} \\
gte-Qwen2-1.5B-instruct & .913 & .569 & .574 & .014 & .745 & .954 & .258 & .670 & .587 \\
Qwen3-Embedding-4B & .943 & .620 & .690 & .110 & .716 & .868 & .257 & .868 & .634 \\
Qwen3-Embedding-8B & .944 & .620 & .770 & .118 & .849 & .893 & .275 & .903 & .672 \\
gte-Qwen2-7B-instruct & .923 & .588 & .642 & .017 & .795 & .945 & .325 & .692 & .616 \\
LLM2Vec-Mistral-7B & .955 & .619 & .812 & .162 & .944 & .902 & .318 & .904 & .702 \\
\midrule
\rowcolor{blue!6}
\multicolumn{10}{l}{\textit{\methodname{} framework (Qwen2.5-1.5B backbone)}} \\
Zero-shot frozen decoder & .956 & .655 & .769 & .152 & .860 & .891 & .311 & .910 & .688 \\
\tunestage{} only & .963 & .682 & .773 & .185 & .884 & .904 & .380 & .939 & .714 \\
\rowcolor{blue!6}
\modelstage{} $\rightarrow$ \tunestage{} & \textbf{.969} & .691 & .804 & .198 & .946 & .934 & .401 & .943 & .736 \\
\rowcolor{blue!6}
\modelstage{} + \matchstage{} $\rightarrow$ \tunestage{} & \textbf{.969} & .695 & .797 & .211 & .959 & .949 & .413 & .950 & .743 \\
\rowcolor{blue!6}
Field-permuted \modelstage{} $\rightarrow$ \tunestage{} & .963 & .697 & \textbf{.817} & \textbf{.224} & .960 & \textbf{.970} & .424 & \textbf{.969} & \textbf{.753} \\
\rowcolor{blue!6}
Field-permuted \modelstage{} + \matchstage{} $\rightarrow$ \tunestage{} (robustness variant) & .968 & .691 & .806 & .219 & \textbf{.972} & .965 & \textbf{.434} & .967 & \textbf{.753} \\
\bottomrule
\end{tabular}}
\vspace{-.15in}
\end{table*}

\myparagraph{Numerical completion.}
Tab.~\ref{tab:imputation-completion-1p5b} evaluates numerical generation through masked-value completion. For each test example, we truncate the serialized input immediately after a selected \texttt{<key: } marker and ask the decoder to generate the corresponding value. A prediction is correct when the parsed value is within 10\% of the held-out value. We use only Stage-1 checkpoints instead of Stage-2 because the task measures general numerical completion before downstream target specialization. Our Stage-1 raises raises mean masked-value completion from 0.106 to 0.448.
Adding \matchstage{} does not further improve numerical completion: Gen-8 slightly reduces to 0.445 with field-order permutation. This is consistent with its different objective: same-record consistency favors representations stable across partial views (Appx. Tab.~\ref{tab:field-dropout}), whereas exact completion requires retaining view-specific numerical details. \matchstage{}'s benefit instead appears in robustness to partial views (Fig.~\ref{fig:matched}(b)).
Overall, Tab.~\ref{tab:imputation-completion-1p5b}; combined with the magnitude-sensitivity and cross-field-context analyses in Sec.~\ref{sec:distinct-capabilities}, they provide behavioral evidence that our Stage-1 improves numerical understanding.


\begin{table*}[t]
\vspace{-.2in}
\caption{\textbf{Stage-1 training improves numerical completion.} We test the masked-value completion by asking the LM to generate the missing value. \textit{Gen-8} is the unweighted mean across eight datasets.
\modelstage{} provides the largest improvement in numerical completion. Adding \matchstage{} does not improve numerical completion, instead it benefits to robustness to partial views (Fig.~\ref{fig:matched}(b)) and end-to-end target prediction (Tab.~\ref{tab:embedding-per-dataset}).
}
\label{tab:imputation-completion-1p5b}
\centering
\scriptsize
\setlength{\tabcolsep}{3.1pt}
\renewcommand{\arraystretch}{1.05}
\resizebox{\textwidth}{!}{%
\begin{tabular}{lrrrrrrrrr}
\toprule
\multirow{2}{*}{Model / recipe}
& \multicolumn{8}{c}{Within-10\% completion accuracy $\uparrow$}
& \multirow{2}{*}{Gen-8 $\uparrow$} \\
\cmidrule(lr){2-9}
& MiniBooNE & HIGGS & Supercond. & YearPred & Naval & CPU act.
& Protein & USDA & \\
\midrule
\multicolumn{10}{l}{\textit{Zero-shot decoder LMs}} \\
Qwen3-4B & .006 & .130 & .292 & .024 & .144 & .194 & .000 & .260 & .131 \\
Qwen3-8B & .004 & .142 & .282 & .024 & .152 & .168 & .000 & .258 & .129 \\
Qwen3.5-9B & .006 & .172 & .280 & .034 & .122 & .174 & .000 & .286 & .134 \\
Qwen3-14B & .002 & .158 & .294 & .030 & .140 & .172 & .000 & .290 & .136 \\
OLMo-2-7B & .002 & .124 & .230 & .006 & .130 & .202 & .000 & .194 & .111 \\
DeepSeek-V2-Lite (16B MoE) & .004 & .132 & .246 & .024 & .156 & .148 & .000 & .220 & .116 \\
gpt-oss-20b & .008 & .128 & .128 & .010 & .008 & .196 & .000 & .144 & .078 \\
Gemma-4-12B & .004 & .132 & .262 & .028 & .184 & .176 & .002 & .284 & .134 \\
Llama-3.1-8B & .002 & .106 & .284 & .008 & .098 & .212 & .000 & .258 & .121 \\
\midrule
\rowcolor{blue!6}
\multicolumn{10}{l}{\textit{\methodname{} framework (Qwen2.5-1.5B backbone)}} \\
Zero-shot frozen decoder & .004 & .126 & .250 & .028 & .086 & .158 & .000 & .196 & .106 \\
\rowcolor{blue!6}
Field-permuted \modelstage{} & \textbf{.318} & \textbf{.292} & \textbf{.690} & \textbf{.062} & .904 & .364 & \textbf{.592} & \textbf{.358} & \textbf{.448} \\
\rowcolor{blue!6}
Field-permuted \modelstage{} + \matchstage{} & .306 & .286 & .676 & \textbf{.062} & \textbf{.912} & \textbf{.382} & .582 & .356 & .445 \\
\bottomrule
\end{tabular}}
\vspace{-.15in}
\end{table*}

\subsubsection{MODEL, MATCH, AND TUNE CONTRIBUTE DISTINCT CAPABILITIES}
\label{sec:distinct-capabilities}

\myparagraph{\modelstage{} increases reliance on the correct cross-field context.}
Tab.~\ref{tab:cross-field-context} evaluates whether the model can correctly do the masked-value completion with correct and mismatched records on three dataset (USDA, Superconductivity, and MiniBooNE). The \textit{correct records} contains the correct values from the same example, while the \textit{mismatched records} contains a mismatched value that preserves the field names, order, and target but replaces with values from another example.
The resulting context gain, correct minus mismatched accuracy $\Delta_{\mathrm{ctx}}$, rises from 0.058 for the frozen model to 0.167 after fixed-order \modelstage{} (Appx.~\ref{app:metrics}), and to 0.271 with field-permuted \modelstage{}.
For field-permuted \modelstage{}, paired record-bootstrap intervals are positive on USDA (0.244, 95\% CI [0.201, 0.289]), Superconductivity (0.472, [0.423, 0.520]), and MiniBooNE (0.096, [0.051, 0.139]).
The context gain $\Delta_{\mathrm{ctx}}$ is positive on all three datasets for both \modelstage{} variants. Thus, across these three diagnostics, \modelstage{} makes masked-value prediction depend more strongly on values from the correct cross-field context, indicating that the model has learned relationships among fields.

\myparagraph{\modelstage{} increases sensitivity to numerical magnitude.}
Changing a leading digit typically alters a number much more than changing a later digit, so a numerically sensitive representation should respond more strongly to the former. We test this by comparing the embedding changes caused by first- and fourth-digit edits. The sensitivity ratio rises from $2.08\times$ for the frozen model to $13.83\times$ with \modelstage{} and $17.82\times$ with field-permuted \modelstage{} (Fig.~\ref{fig:place-value}(b); Appx. Fig.~\ref{fig:digit-sensitivity}). Relative to the frozen model, \modelstage{} increases the ratio by 11.75 (95\% CI [7.04, 17.09]) under a paired record bootstrap. Field permutation adds 3.99 relative to \modelstage{}, but its interval crosses zero (95\% CI [$-3.91$, 13.17]). Thus, \modelstage{} increases sensitivity to numerically consequential digit changes, while this diagnostic does not resolve an additional effect of field permutation.

\begin{table*}[t]
\vspace{-.2in}
\caption{\textbf{\modelstage{} improves the utilization of cross-field context.}
For each masked field, the correct and mismatched prompts have identical field names, order, target, and scorer; only the values in mismatched prompts are replaced by values from another record.
$\Delta_{\mathrm{ctx}}$ is correct minus mismatched within-10\% accuracy. The first three columns are macro averages over three dataaset; the remaining columns show the corresponding per-dataset context gains $\Delta_{\mathrm{ctx}}$. }
\label{tab:cross-field-context}
\centering
\scriptsize
\setlength{\tabcolsep}{5pt}
\renewcommand{\arraystretch}{1.08}
\begin{tabular}{lrrrrrr}
\toprule
\multirow{2}{*}{Model / recipe}
& \multicolumn{3}{c}{Macro within-10\% accuracy}
& \multicolumn{3}{c}{Per-dataset $\Delta_{\mathrm{ctx}}\uparrow$} \\
\cmidrule(lr){2-4}\cmidrule(lr){5-7}
& Correct $\uparrow$ & Mismatched & $\Delta_{\mathrm{ctx}}\uparrow$ & USDA & Supercond. & MiniBooNE \\
\midrule
Frozen base & .142 & .084 & +.058 & +.064 & +.115 & -.004 \\
\modelstage{} & .338 & .171 & +.167 & +.170 & +.284 & +.047 \\
Field-permuted \modelstage{} & \textbf{.463} & .193 & \textbf{+.271} & \textbf{+.244} & \textbf{+.472} & \textbf{+.096} \\
\bottomrule
\end{tabular}
\end{table*}

\begin{figure}[t]
\vspace{-.1in}
    \centering
    \includegraphics[width=\textwidth]{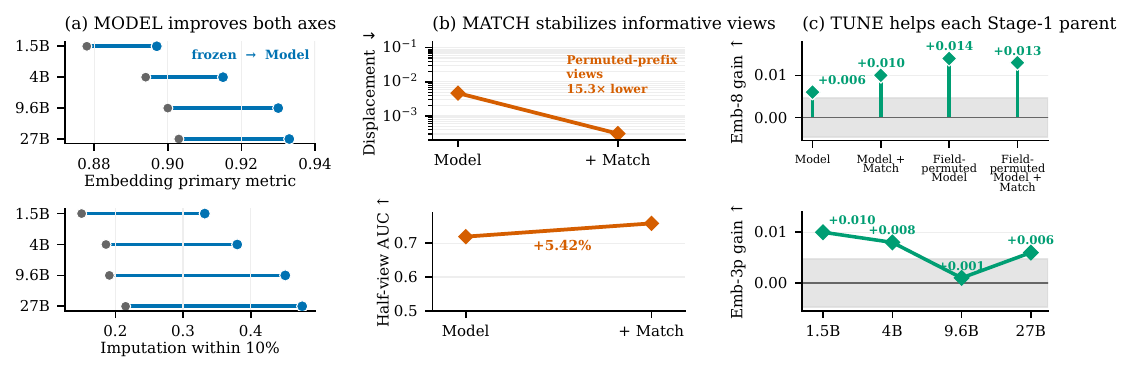}
    \vspace{-.3in}
    \caption{\textbf{\modelstage{} and \tunestage{} improve their intended capabilities, while
    \matchstage{} stabilizes informative partial views.} (a) Across three-dataset, fixed-order \modelstage{} improves both the primary target-prediction metric and numerical completion relative to the frozen checkpoint at every scale. (b) For permuted-prefix views, \matchstage{} reduces embedding displacement by $15.3\times$ and raises half-view AUC by 5.4\%. (c) The upper panel shows that \tunestage{} improves \textsc{Emb-8} for every tested Stage-1 parent at 1.5B; the lower panel shows a positive three-dataset gain through 27B for the field-permuted \modelstage{}+\matchstage{} parent.
    }
    \vspace{-.15in}
    \label{fig:matched}
\end{figure}

\myparagraph{\matchstage{} improves robustness to missing fields.} Missing fields can shift a record representation and degrade its prediction. We test whether \matchstage{} reduces this sensitivity by removing half of the input record fields and evaluating prediction from the same incomplete views with and without \matchstage{}. When added to field-permuted \modelstage{}, \matchstage{} reduces the full-to-incomplete representation displacement from 0.00468 to 0.000306 ($15.3\times$, 95\% CI [$15.257\times$, $15.354\times$]) and raises mean incomplete-view ROC--AUC by 5.4\% from 0.719 to 0.758 (95\% CI [0.023, 0.056])
(Fig.~\ref{fig:matched}(b); Appx. Fig.~\ref{fig:match-robustness} and Tab.~\ref{tab:field-dropout}). 
Thus, when both views retain useful target information, \matchstage{} improves prediction under missing fields rather than complete-record accuracy.

\myparagraph{\matchstage{} can hurt when paired views contain complementary predictive information.} For disjoint-half views, adding \matchstage{} to fixed-order \modelstage{} reduces representation displacement from 0.0916 to 0.0120, but lowers mean half-view AUC from 0.584 to 0.533 (Appx. Tab.~\ref{tab:field-dropout}). Each half may contain predictive fields absent from the other, so enforcing consistency can suppress view-specific but task-relevant information. \matchstage{} is therefore most appropriate when paired views preserve compatible predictive evidence. 

\myparagraph{\tunestage{} uses labeled examples to improve target prediction.}
In Tab.~\ref{tab:embedding-per-dataset}, at 1.5B, adding \tunestage{} raises \textsc{Emb-8} by 0.8--1.9\% relative to its Stage-1 parent: fixed-order \modelstage{}, fixed-order \modelstage{}+\matchstage{}, field-permuted \modelstage{}, and field-permuted \modelstage{}+\matchstage{} (Fig.~\ref{fig:matched}(c); Appx. Tab.~\ref{tab:aggregate-ablation-controls}). In Tab.~\ref{tab:model-size-scaling-primary}, compared with \tunestage{} alone, our \methodname{} improves the target-prediction score at every tested scale from 1.5B to 27B, with relative gains of $+2.6\%$, $+2.2\%$, $+1.0\%$, and $+1.6\%$, respectively.
In contrast, masked-value imputation decreases for every tested parent. Thus, labeled \tunestage{} examples specialize the Stage-1 representation for target prediction while giving up some general record-completion ability.

\begin{figure}[t]
    \centering
    \includegraphics[width=\textwidth]{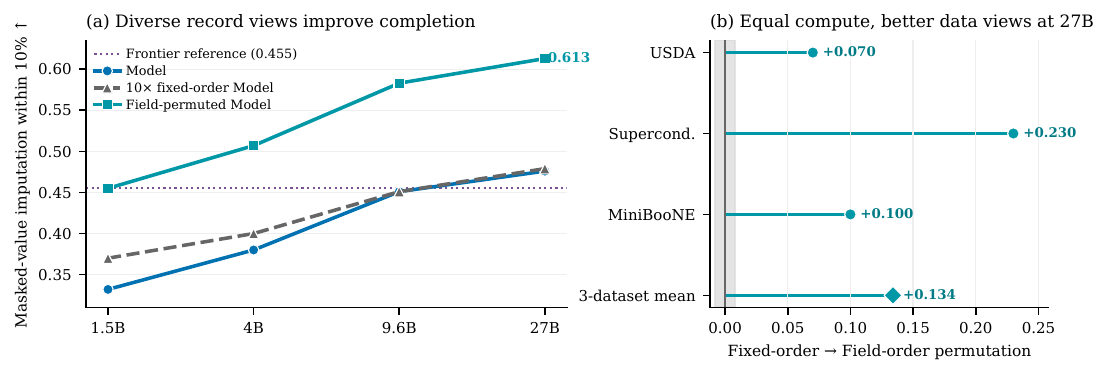}
    \vspace{-.3in}
    \caption{\textbf{Field-order permutation improves completion beyond an equal-update
    fixed-order control.} (a) Mean within-10\% masked-value imputation over three-dataset from 1.5B to 27B, comparing standard fixed-order \modelstage{},
    extended fixed-order training, and field-permuted \modelstage{};
    the dotted line marks the 0.455 frontier-model reference used in this diagnostic. (b) At 27B, field-permuted \modelstage{} improves every dataset over the equal-update fixed-order control, increasing the three-dataset mean by 0.134; the gray band marks the $\pm0.008$
    same-configuration imputation spread.
    }
    \label{fig:trajectory}
    \vspace{-.2in}
\end{figure}

\myparagraph{Field-order permutation helps \modelstage{} learn beyond fixed field positions.}
At 1.5B, field-order permutation raises the digit-sensitivity ratio from $13.83\times$ to $17.82\times$ (Fig.~\ref{fig:place-value}(b)), target prediction emb-8 increase from 0.743 to 0.753 (Tab.~\ref{tab:embedding-per-dataset}), masked-value imputation from 0.401 to 0.448 (Tabs.~\ref{tab:imputation-completion-1p5b}), and context gain from 0.167 to 0.271 when comparing fixed-order with field-permuted \modelstage{}
(Tab.~\ref{tab:cross-field-context}). The equal-update 27B shares the same pattern: field-permuted \modelstage{} exceeds the fixed-order control on every primary dataset and raises mean imputation from 0.479 to 0.613 and the three-dataset target-prediction mean from 0.920 to 0.936 (Fig.~\ref{fig:trajectory}).
Together, these comparisons support field-order permutation as a data augmentation that improves numerical sensitivity, target prediction, completion, and context use.

\subsection{Ablation Studies}
\label{sec:ablations}


\myparagraph{Our two-stage training framework gains persist from model size 1.5B to 27B.} Tab.~\ref{tab:model-size-scaling-primary} compares target-prediction scores across the 1.5B, 4B, 9.6B, and 27B checkpoints. Fixed-order \modelstage{} consistently outperform its corresponding frozen checkpoint  (2.2--3.3\%), while the field-permuted \modelstage{}+\matchstage{}$\rightarrow$\tunestage{} variant outperforms \tunestage{} alone (1.0--2.6\%) across all four sizes.
Adding \tunestage{} to \modelstage{}+\matchstage{} also yields consistent gains (0.1--1.1\%). Appx. Fig.~\ref{fig:scaling} and Tab.~\ref{tab:model-size-imputation-full} report masked-value imputation results, and Appx. Tab.~\ref{tab:complete-27b-grid} provides the complete 27B recipe grid.


\begin{table}[t]
\caption{
\textbf{Target-prediction comparisons across the tested model sizes.}
Each cell reports before$\rightarrow$after and the absolute difference in parentheses for the mean of MiniBooNE ROC--AUC and Superconductivity and USDA $R^2$. All three comparisons show positive point differences at the four tested model sizes.
}
\label{tab:model-size-scaling-primary}
\centering
\scriptsize
\setlength{\tabcolsep}{3.8pt}
\renewcommand{\arraystretch}{1.06}
\begin{tabular}{lccc}
\toprule
  Scale &
  \shortstack{Frozen checkpoint $\rightarrow$ \modelstage{} } &
  \shortstack{Field-permuted \modelstage{}+\matchstage{} $\rightarrow$ \methodname{}} &
  \shortstack{\tunestage{} only $\rightarrow$ \methodname{}} \\
  \midrule
1.5B & .878$\rightarrow$.897 (+.019) & .904$\rightarrow$.914 (+.010) & .891$\rightarrow$.914 (+.023) \\
4B   & .894$\rightarrow$.915 (+.021) & .922$\rightarrow$.930 (+.008) & .910$\rightarrow$.930 (+.020) \\
9.6B & .900$\rightarrow$.930 (+.030) & .934$\rightarrow$.935 (+.001) & .926$\rightarrow$.935 (+.009) \\
27B  & .903$\rightarrow$.933 (+.030) & .936$\rightarrow$.942 (+.006) & .927$\rightarrow$.942 (+.015) \\
\bottomrule
\end{tabular}
\vspace{-.15in}
\end{table}

\myparagraph{Generality to text and categorical fields.}
Appx.~\ref{app:mixed-type} shows that our two-stage training framework also generalizes to mixed text and numerical data. Evaluations on two non-numerical datasets (up to 95.3\% non-numeric) show trends consistent with those in Tab.~\ref{tab:embedding-per-dataset}.

\myparagraph{Sensitivity to loss weights.}
Appx.~\ref{app:loss-weight-sensitivity} varies the weighting in Eq.~\ref{eq:adapt}. The default \modelstage{}--\matchstage{} weighting achieves the highest $R^2$, and all tested $\lambda_2\in[0.5,2.0]$ settings are within 0.005.

\section{Conclusion}
\label{sec:conclusion}
Numerical records require language models to represent both magnitude within values and relationships across fields. Our results support a hierarchical division of labor in \methodname{}. \modelstage{} is the foundation: it adapts the decoder to numerical-record structure and improves both target prediction and numerical completion. \tunestage{} then specializes the resulting representation for downstream target prediction, while reducing general record-completion ability. \matchstage{} is optional: it can stabilize informative partial views, but it does not increase complete-record performance in the strongest field-permuted comparison and can reduce prediction quality when paired views contain complementary information. We therefore recommend \modelstage{}$\rightarrow$\tunestage{} as the core path, with \matchstage{} added only when partial-view robustness is required and the intended view construction has been validated.


\label{main-end}

\subsection*{AI use statement}
Generative AI tools were used for literature organization, code assistance, experiment
bookkeeping, and polishing this manuscript. All citations were checked
against source records; all numerical results were produced by executed code and checked against stored evaluation artifacts and logs. The authors reviewed all AI-assisted work and take responsibility for the final text, claims, code, and results.

\subsection*{Ethics statement}
This study presents minimal direct ethical risk. It uses public scientific and numerical benchmark datasets, involves no human participants or intervention, and does not collect personal or sensitive data. All evaluations are conducted offline for research purposes. Any future use in consequential applications would require separate domain-specific validation and oversight.

\subsection*{Reproducibility statement}
Appx.~\ref{app:repro} documents dataset construction, splits, serialization, and target exclusion, while Appx.~\ref{app:metrics} defines each evaluation task, metric, baseline, and aggregate. Appx.~\ref{app:model-architectures}, \ref{app:arms}, and~\ref{app:training-details} specify the checkpoint families, complete recipe registry, optimization settings, seeds, and view construction. Full per-dataset results and matched controls appear in Appx.~\ref{app:perdataset}; mechanism and failure-analysis protocols appear in Appx.~\ref{app:mechanism}; and complete model-size results appear in Appx.~\ref{app:model-size-results}. Appx.~\ref{app:secondary-decoder} records the direct-decoder prompts, token limits, parsing rules, and excluded invalid cases. The supplementary material includes the result files and scripts used to regenerate the reported tables.

\bibliography{iclr2027_conference}
\bibliographystyle{iclr2027_conference}

\clearpage
\appendix

\section{Limitations}

The model-size sweep changes architecture and tokenizer vocabulary and these experiments therefore do not establish a fitted scaling law.
Appx.~\ref{app:mixed-type} does extend the full recipe block to two datasets whose records are largely categorical, including one in which 95.3\% of the value characters are non-numeric, and there no recipe falls below the frozen baseline; that dataset is however close to a frozen probe's ceiling, and the main results remain almost entirely numerical, so this evidence bounds the mixed-type question rather than settling it. Finally, \tunestage{} directly observes the target while reducing imputation, so our conclusions are limited to post-training numerical records and target specialization rather than broad tabular or label-free generality.


\section{Intuition: LLM Numeracy Issue}
\label{app:digit-blindness}

A decimal reaches the model as several tokens, and current models split it in one of two ways. Under both
conventions, changing a single digit changes exactly one token, so a one-hot training target charges the
same price whether that digit was worth $23.4\%$ or $0.0234\%$ of the value.

\begin{itemize}
    \item \textbf{One digit per token.} Every digit becomes its own token, so \texttt{0.4271} is six
    tokens, (\texttt{0})~(\texttt{.})~(\texttt{4})~(\texttt{2})~(\texttt{7})~(\texttt{1}); used by Qwen2.5/3/3.5,
    Mistral, Gemma-4, DeepSeek-V2 and Llama-2.
    \item \textbf{Groups of up to three digits.} Digits are merged into chunks. In Fig.~\ref{fig:place-value}(a),
    \texttt{0.427132} is displayed as four tokens,
    (\texttt{0})~(\texttt{.})~(\texttt{427})~(\texttt{132}). This convention is used by GPT-3.5 and GPT-4
    (\texttt{cl100k\_base}), GPT-4o and the o-series (\texttt{o200k\_base}), Llama-3, OLMo-2 and phi-4.
\end{itemize}

We call a model \emph{digit-blind} when its record embedding moves by about the same amount regardless of
which digit was edited. We measure this by editing either the first or the fourth significant digit. Then compute $\bar{d}_1/\bar{d}_4$: how far the embedding moves when breaking the most important digit ($\bar{d}_1$), divided by how far it moves when breaking the least important one ($\bar{d}_4$).
For example, number
\texttt{0.4271}, editing the first digit gives \texttt{0.5271} and moves the value by
$23.4\%$, while editing the fourth gives \texttt{0.4272} and moves it by $0.0234\%$; each edit replaces
exactly one token. The $1000\times$ contrast is only a numerical input-scale reference. The diagnostic
asks whether first-digit edits produce stronger embedding separation than fourth-digit edits, not whether
the embedding-distance ratio reaches $1000\times$. Tab.~\ref{tab:digit-blindness} reports this for seven frozen pretrained models spanning both
conventions, using the protocol of Appx.~\ref{app:mechanism}.

Across the seven surveyed models, ratios lie between $1.12$ and $2.14$ against an input contrast of
$1000\times$, and the two conventions overlap completely. On this diagnostic, digit-blindness therefore
appears across the surveyed pretrained decoders rather than being isolated to one tokenizer or one model
family, which motivates addressing it through training rather than re-tokenizing.

\begin{table}[H]
\caption{Digit-blindness in frozen pretrained models, under both tokenization conventions. Each model is
evaluated with its own tokenizer, so the accepted record set is derived per model and the sets are not
identical across rows; the Qwen2.5-1.5B row is therefore an independent re-derivation of the frozen row of
Tab.~\ref{tab:digit-sensitivity} ($2.14$ here versus $2.08$ there). Larger ratios indicate greater
leading-digit sensitivity under this proxy. The reference row contrasts a first-digit edit (\texttt{0.4271}$\to$\texttt{0.5271}, a $23.4\%$
change in value) with a fourth-digit edit (\texttt{0.4271}$\to$\texttt{0.4272}, $0.0234\%$), a ratio
of $1000\times$; it is a reference contrast, not an expected embedding ratio. MiniBooNE, 200 accepted records per model, frozen checkpoints only.}
\label{tab:digit-blindness}
\centering
\small
\begin{tabular}{llrr}
\toprule
Tokenization & Model & $\bar{d}_1/\bar{d}_4$ $\uparrow$ & Spearman $\rho$ $\uparrow$ \\
\midrule
One digit per token & Qwen2.5-1.5B-Instruct & 2.14 & 0.276 \\
One digit per token & Qwen3-14B & 2.04 & 0.196 \\
One digit per token & Mistral-7B-v0.3 & 1.65 & 0.191 \\
One digit per token & Gemma-4-12B & 1.12 & 0.131 \\
\midrule
Groups of up to three & phi-4 & 1.94 & 0.353 \\
Groups of up to three & OLMo-2-7B & 1.81 & 0.159 \\
Groups of up to three & Llama-3.1-8B & 1.51 & 0.238 \\
\midrule
-- & Underlying value change (e.g.\ \texttt{0.4271}$\to$\texttt{0.5271}) & 1000.00 & -- \\
\bottomrule
\end{tabular}
\end{table}

\section{Dataset and Evaluation Metrics Details}
\label{app:repro}

\subsection{Dataset construction}

Each OpenML dataset is converted to a float feature matrix, capped at 30,000 records where
necessary, randomly permuted with NumPy seed 42, and split 80/10/10. Values are formatted with four
significant digits; integer-valued floats are written as integers. Keys remain in the metadata
column order. The materialized split is reused by every checkpoint, probe, and generation test.

The benchmark is deliberately numerical. MiniBooNE, HIGGS, Superconductivity, CPU activity,
Naval propulsion, Protein structure, and YearPredictionMSD contain numerical input fields only.
USDA FoodData contains 48 numerical nutrient fields and one natural-language food description.
Thus 339 of 340 schema fields are numerical, and 95.3\% of records come from numerical-only
datasets. We do not synthesize descriptions or convert numbers into explanatory prose.

\begin{table}[H]
\caption{Public evaluation corpus. Seven of eight datasets are numerical-only; across schemas,
339/340 input fields are numerical. ``Records'' includes train, validation, and test splits.}
\label{tab:datasets}
\centering
\small
\begin{tabular}{lcrrrl}
\toprule
Dataset & Task & Numeric fields & Text fields & Records & Target \\
\midrule
MiniBooNE & cls. & 50 & 0 & 30,000 & signal \\
HIGGS & cls. & 28 & 0 & 30,000 & class \\
Superconductivity & reg. & 79 & 0 & 21,263 & critical temperature \\
USDA FoodData & reg. & 48 & 1 & 7,896 & energy (kcal) \\
CPU activity & reg. & 21 & 0 & 8,192 & user CPU \% \\
Naval propulsion & reg. & 14 & 0 & 11,934 & compressor decay \\
Protein structure & reg. & 9 & 0 & 30,000 & RMSD \\
YearPredictionMSD & reg. & 90 & 0 & 30,000 & year \\
\midrule
\multicolumn{2}{l}{Total} & 339 & 1 & 169,285 & 7/8 numeric-only \\
\bottomrule
\end{tabular}
\end{table}

The downstream target is stored separately and is absent from every Stage-1 record-adaptation
serialization. \tunestage{} adds
the target only through a question--answer suffix, as illustrated in Fig.~\ref{fig:examples}. Under
field-order permutation, the values and field set stay fixed; only the order of \kv{} units changes.
After truncation to the model's token budget, however, different field orders can expose different
visible field subsets; this partial-view diversity is the intervention tested in the ablation.
Tab.~\ref{tab:dataset-full} gives all split sizes, target ranges, and generation sample counts.

\begin{table}[H]
\caption{Complete dataset statistics and input modality. ``Gen. $n$'' is the number of test records
used for masked-value imputation and field lookup for each checkpoint.}
\label{tab:dataset-full}
\centering
\small
\resizebox{\textwidth}{!}{%
\begin{tabular}{lrrrrrrl}
\toprule
Dataset & Numeric & Text & Train & Val. & Test & Gen. $n$ & Target range / classes \\
\midrule
MiniBooNE & 50 & 0 & 24,000 & 3,000 & 3,000 & 500 & binary \\
HIGGS & 28 & 0 & 24,000 & 3,000 & 3,000 & 500 & binary \\
Superconductivity & 79 & 0 & 17,010 & 2,126 & 2,127 & 500 & $[0.00021,185]$ \\
USDA FoodData & 48 & 1 & 6,316 & 789 & 791 & 500 & $[0,902]$ \\
CPU activity & 21 & 0 & 6,553 & 819 & 820 & 500 & $[0,99]$ \\
Naval propulsion & 14 & 0 & 9,547 & 1,193 & 1,194 & 500 & $[0.95,1.0]$ \\
Protein structure & 9 & 0 & 24,000 & 3,000 & 3,000 & 500 & $[0,20.999]$ \\
YearPredictionMSD & 90 & 0 & 24,000 & 3,000 & 3,000 & 500 & $[1922,2010]$ \\
\bottomrule
\end{tabular}}
\end{table}

\subsection{Evaluation dimensions and metrics}
\label{app:metrics}

We organize evaluation along two complementary dimensions. The \emph{embedding axis} measures
what target information is linearly recoverable from frozen record representations. The
\emph{generation axis} measures whether the decoder can directly produce or score numerical record
content without a fitted readout. Geometry diagnostics explain changes in representation space and
are not folded into any headline aggregate.

\paragraph{Embedding axis.}
The embedding axis evaluates each trained checkpoint through post-hoc supervised probes. After
training for an arm is complete, we freeze the language model and extract one representation from
each record with the target omitted. We then use scikit-learn to fit a separate linear readout on the
task's training split: logistic regression for classification and Ridge regression for regression.
We select regularization on validation, refit the selected probe on train${+}$validation, and score
it once on test. A \emph{probe prediction} is the output of this readout; the probe neither updates
the decoder nor participates in continued pretraining or \tunestage{}.
\begin{equation}
\begin{aligned}
\widetilde z &= \operatorname{Standardize}_{\mathrm{train}}\!\left(z_\theta(x)\right),\\
p_\phi(y=1\mid x) &= \sigma\!\left(w_c^\top\widetilde z+b_c\right),
& \widehat y_\psi(x) &= w_r^\top\widetilde z+b_r .
\end{aligned}
\label{eq:embedding-probes}
\end{equation}
Here $z_\theta(x)$ is the record representation produced by the frozen trained checkpoint,
$\phi=(w_c,b_c)$ is the logistic-regression probe, and $\psi=(w_r,b_r)$ is the Ridge-regression
probe. These are separate scikit-learn models, not an MLP or an added LM layer.

In all examples below, italic labels describe the evaluation pipeline and are not part of the model
input. Monospaced lines show the input text; $\ldots$ marks fields omitted only for display. Scoring
references are labelled separately: they are absent from prediction and imputation inputs, whereas
field lookup deliberately leaves the requested value in the record. Because predictions vary by
checkpoint, the examples state the returned object or parsed prediction rather than substituting the
reference value as if it were an observed model output.

\begin{itemize}[leftmargin=1.35em,itemsep=6pt,topsep=3pt]
    \item \textbf{Classification probe.} For MiniBooNE and HIGGS, logistic regression maps the
    frozen vector to a class probability. We choose
    $C\in\{0.01,0.1,1,10\}$ by validation ROC-AUC. Test metrics are accuracy
    $N^{-1}\sum_i\mathbf{1}[\widehat y_i=y_i]$ and ROC-AUC, which measures whether positive
    examples receive higher scores than negative examples across thresholds.

    \item \textbf{Regression probe.} For the six continuous-target datasets, Ridge regression
    maps the frozen vector to a scalar. We choose
    $\alpha\in\{0.1,1,10,100,1000\}$ by validation $R^2$. Test metrics are
    \begin{equation*}
    R^2=1-\frac{\sum_i(y_i-\widehat y_i)^2}{\sum_i(y_i-\bar y)^2},\qquad
    \mathrm{MAE}=\frac{1}{N}\sum_i|y_i-\widehat y_i|,
    \end{equation*}
    together with Pearson's $r$ between $y$ and $\widehat y$. Higher $R^2$ and Pearson $r$ are
    better; lower MAE is better.

\end{itemize}

\medskip
\noindent
Four worked examples follow, one per probe type and per record composition. Examples~1 and~3 come
from the purely numerical benchmark; Examples~2 and~4 come from the mixed-type datasets of
App.~\ref{app:mixed-type}, where most of a record is words rather than digits. In each case the
upper panel is exactly the string handed to the frozen decoder and the lower panel states what the
probe returns and what it is scored against.
\label{app:probe-cases}

\begin{probecase}{MiniBooNE --- classification probe (purely numerical record)}
\begin{probepanel}{caseinput}{Direct LM input (record abridged; label absent)}
\texttt{<ParticleID\_0: 4.192> \ldots <ParticleID\_49: 0.2366>}
\end{probepanel}
\begin{probepanel}{caseprobe}{Probe and scoring}
\textit{Representation passed to scikit-learn:} standardized vector $\widetilde z$.\\
\textit{Probe output:} probability $p_\phi(y=1\mid x)$; the evaluator predicts
$\widehat y=\mathbf{1}[p_\phi(y=1\mid x)\geq0.5]$.\\
\textit{Gold target, used only for scoring:} $y=1$.
\end{probepanel}
\end{probecase}

\begin{probecase}{Census-KDD --- classification probe (95.3\% of value characters non-numeric)}
\begin{probepanel}{casemixed}{Direct LM input (record abridged; label \texttt{V42} absent)}
\texttt{<V1: 65> <V2: Not in universe> <V5: 12th grade no diploma>}\\
\texttt{<V8: Married-civilian spouse present> <V11: Black> <V13: Female> \ldots}\\
\texttt{<V35: United-States> <V36: Native- Born in the United States> <V41: 94>}
\end{probepanel}
\begin{probepanel}{caseprobe}{Probe and scoring}
\textit{Representation passed to scikit-learn:} standardized vector $\widetilde z$.\\
\textit{Probe output:} probability $p_\phi(y=1\mid x)$; the evaluator predicts
$\widehat y=\mathbf{1}[p_\phi(y=1\mid x)\geq0.5]$.\\
\textit{Gold target, used only for scoring:} $y=0$, the \texttt{- 50000.} income class.
\end{probepanel}
\end{probecase}

\begin{probecase}{Superconductivity --- regression probe (purely numerical record)}
\begin{probepanel}{caseinput}{Direct LM input (record abridged; target \texttt{criticaltemp} absent)}
\texttt{<mean\_atomic\_mass: 88.94> \ldots <wtd\_std\_Valence: 0.4432>}
\end{probepanel}
\begin{probepanel}{caseprobe}{Probe and scoring}
\textit{Representation passed to scikit-learn:} standardized vector $\widetilde z$.\\
\textit{Probe output:} checkpoint-dependent scalar $\widehat y_\psi(x)$.\\
\textit{Gold target, used only for scoring:} $y=9.5$.
\end{probepanel}
\end{probecase}

\begin{probecase}{NYC-Taxi --- regression probe (3.0\% of value characters non-numeric)}
\begin{probepanel}{casemixed}{Direct LM input (record abridged; target \texttt{tip\_amount} absent)}
\texttt{<VendorID: 2> <store\_and\_fwd\_flag: N> <RatecodeID: 1> <PULocationID: 97>}\\
\texttt{<DOLocationID: 61> <passenger\_count: 1> <extra: 0> <mta\_tax: 0.5> \ldots}\\
\texttt{<total\_amount: 11.8> <lpep\_dropoff\_datetime\_minute: 5>}
\end{probepanel}
\begin{probepanel}{caseprobe}{Probe and scoring}
\textit{Representation passed to scikit-learn:} standardized vector $\widetilde z$.\\
\textit{Probe output:} checkpoint-dependent scalar $\widehat y_\psi(x)$.\\
\textit{Gold target, used only for scoring:} $y=1.0$.
\end{probepanel}
\end{probecase}

All checkpoints use identical record inputs and data splits. \textsc{Emb-8} is the unweighted mean
of the two classification ROC-AUC values
and six regression $R^2$ values; because these metrics have different semantics, the aggregate is
a compact summary rather than a substitute for per-dataset results.

Embedding diagnostics include effective rank, mean pairwise cosine, metric faithfulness, and CKA.
Effective rank is $\exp(-\sum_j p_j\log p_j)$ for the normalized eigenvalues of the centered
embedding covariance; larger values indicate that variation occupies more directions. Mean
pairwise cosine measures anisotropy; values near one mean records point in nearly the same
direction. Metric faithfulness is the Spearman correlation between cosine distance and absolute
target difference over sampled test pairs. Linear CKA measures similarity between the spaces
produced by two checkpoints.

\paragraph{Generation axis.}
The generation axis evaluates the frozen decoder directly, with no fitted probe. On each dataset we
use 500 held-out records. Masked-value imputation and field lookup use the original 320-token
decoder harness, greedy decoding, at most 12 new tokens, and deterministic numeric parsing; parse
failures count as incorrect. Held-out likelihood uses the same 320-token input cap. Target emission
is a separate, training-matched diagnostic with a 496-token cap and is valid only for the four
datasets whose complete prompt fits that budget. The four tasks isolate target prediction,
missing-value inference, copying, and distribution modeling, respectively.

\begin{itemize}[leftmargin=1.35em,itemsep=6pt,topsep=3pt]

    \item \textbf{Masked-value imputation.} We deterministically choose one field, truncate the
    serialization immediately after its key marker, and ask the model to continue with the missing
    value. The primary metric is within-10\% accuracy:
    \begin{equation*}
    \frac{1}{N}\sum_{i=1}^{N}
    \mathbf{1}\!\left[|\widehat v_i-v_i|\leq0.1\max(|v_i|,10^{-6})\right].
    \end{equation*}
    Exact match after four-significant-figure normalization, MAE, median absolute error, and
    parse-failure rate are secondary metrics. The median floor predicts each field's train-split
    median conditional on field identity.

    Example~\ref{case:impute} shows one such prompt.

\end{itemize}

\begin{probecase}[label=case:impute]{MiniBooNE --- masked-value imputation (no fitted probe)}
\begin{probepanel}{caseinput}{Direct LM prompt (record prefix abridged; masked value absent)}
\texttt{<ParticleID\_0: 4.192> \ldots <ParticleID\_35: }
\end{probepanel}
\begin{probepanel}{caseprobe}{Completion and scoring}
\textit{LM completion:} checkpoint-dependent continuation text.\\
\textit{Parsed prediction:} first number in the continuation, denoted $\widehat v$.\\
\textit{Gold feature value, used only for scoring:} $v=0.3543$.
\end{probepanel}
\end{probecase}

To preserve comparability with the main results, \textsc{Gen-8} is the unweighted mean of the
within-10\% imputation accuracy over the eight datasets. Target-generation results are reported
separately because classification accuracy and regression log-ratio error are not commensurate.

Held-out record perplexity remains under the generation axis because it evaluates the record
distribution rather than an external capability.

\begin{table}[H]
\caption{Dataset-to-metric map. Every dataset uses masked-value imputation, field lookup, and
perplexity; target emission is reported only where the complete training-matched prompt fits the
496-token budget. Embedding metrics depend on whether the target is categorical or continuous.}
\label{tab:metric-map}
\centering
\small
\resizebox{\textwidth}{!}{%
\begin{tabular}{lllll}
\toprule
Dataset & Prediction target & Task type & Embedding metrics & Generation metrics \\
\midrule
MiniBooNE & signal & binary classification & Acc., ROC-AUC $\uparrow$ & impute $\uparrow$; lookup $\uparrow$; PPL $\downarrow$ \\
HIGGS & class & binary classification & Acc., ROC-AUC $\uparrow$ & target Acc. $\uparrow$; impute $\uparrow$; lookup $\uparrow$; PPL $\downarrow$ \\
Superconductivity & critical temperature & regression & $R^2,r\uparrow$; MAE $\downarrow$ & impute $\uparrow$; lookup $\uparrow$; PPL $\downarrow$ \\
USDA FoodData & energy (kcal) & regression & $R^2,r\uparrow$; MAE $\downarrow$ & impute $\uparrow$; lookup $\uparrow$; PPL $\downarrow$ \\
CPU activity & user CPU \% & regression & $R^2,r\uparrow$; MAE $\downarrow$ & target log error $\downarrow$; impute $\uparrow$; lookup $\uparrow$; PPL $\downarrow$ \\
Naval propulsion & compressor decay & regression & $R^2,r\uparrow$; MAE $\downarrow$ & target log error $\downarrow$; impute $\uparrow$; lookup $\uparrow$; PPL $\downarrow$ \\
Protein structure & RMSD & regression & $R^2,r\uparrow$; MAE $\downarrow$ & target log error $\downarrow$; impute $\uparrow$; lookup $\uparrow$; PPL $\downarrow$ \\
YearPredictionMSD & year & regression & $R^2,r\uparrow$; MAE $\downarrow$ & impute $\uparrow$; lookup $\uparrow$; PPL $\downarrow$ \\
\bottomrule
\end{tabular}}
\end{table}

\section{Experiment Details}

\subsection{Model Architectures}
\label{app:model-architectures}

Tab.~\ref{tab:model-architectures} summarizes the checkpoints used in the primary experiments and
model-size robustness sweep. The Qwen3.5 and Qwen3.8 checkpoints use the same hybrid decoder family;
the vision components exposed by these checkpoints are unused, and all experiments operate on the
text decoder only.

\begin{table}[H]
\caption{Model checkpoints used in our experiments. Vocabulary denotes tokenizer vocabulary size.}
\label{tab:model-architectures}
\centering
\small
\resizebox{\textwidth}{!}{%
\begin{tabular}{lclc}
\toprule
Checkpoint & Scale & Decoder architecture & Vocabulary \\
\midrule
Qwen2.5-1.5B-Instruct & 1.5B & Dense Qwen2 decoder & 151,936 \\
Qwen3-4B-Instruct-2507 & 4B & Dense Qwen3 decoder & 151,936 \\
Qwen3.5-9B & 9.6B & 32-layer hybrid full/linear-attention decoder & 248,320 \\
Qwen3.8-27B & 27B & 64-layer hybrid full/linear-attention decoder & 248,320 \\
\bottomrule
\end{tabular}}
\end{table}

\subsection{Complete Arm Registry}
\label{app:arms}

\begin{table}[H]
\caption{The principal arms. ``Views'' is the number of deterministic row serializations per epoch.
All \modelstage{} recipes use weight 0.5 for that loss. Every serialization is right-truncated at
512 tokens; a ``permuted-prefix view'' denotes independent field-order permutation followed by truncation. \recipedefs}
\centering
\small
\resizebox{\textwidth}{!}{%
\begin{tabular}{llllll}
\toprule
Recipe & Stage-1 order / views & Stage-1 objective & Stage 2 & Budget & Purpose \\
\midrule
Frozen decoder & fixed-order / 1 & none & -- & -- & frozen initialization \\
\modelstage{} & fixed-order / 1 & \modelstage{} & -- & 2 epochs & primary baseline \\
\matchstage{} only & fixed-order / 1 & \matchstage{} only (disjoint views) & -- & 2 epochs & consistency-only control \\
\modelstage{} + \matchstage{} & fixed-order / 1 & \modelstage{} + \matchstage{} (disjoint views) & -- & 2 epochs & \modelstage{}+\matchstage{} recipe \\
\tunestage{} only & fixed-order / 1 & none & fixed-order \tunestage{} & \tunestage{} only & no-Stage-1 control \\
Extended-compute \modelstage{} & fixed-order / 1 & \modelstage{} & -- & 20 epochs & 10$\times$ compute \\
Field-permuted \modelstage{} & field-permuted / 10 & \modelstage{} & -- & 2 epochs & field-permuted baseline \\
Field-permuted \matchstage{} only & field-permuted / 10 & \matchstage{} only (permuted-prefix views) & -- & 2 epochs & consistency-only control \\
Field-permuted \modelstage{} + \matchstage{} & field-permuted / 10 & \modelstage{} + \matchstage{} (permuted-prefix views) & -- & 2 epochs & field-permuted \modelstage{}+\matchstage{} recipe \\
\modelstage{} $\rightarrow$ \tunestage{} & fixed-order / 1 & \modelstage{} & fixed-order \tunestage{} & 2 + 2 epochs & supervised baseline \\
\modelstage{} + \matchstage{} $\rightarrow$ \tunestage{} & fixed-order / 1 & \modelstage{} + \matchstage{} (disjoint views) & fixed-order \tunestage{} & 2 + 2 epochs & supervised \modelstage{}+\matchstage{} recipe \\
Field-permuted \modelstage{} $\rightarrow$ \tunestage{} & field-permuted / 10 & \modelstage{} & fixed-order \tunestage{} & 2 + 2 epochs & supervised control \\
Field-permuted \modelstage{} + \matchstage{} $\rightarrow$ \tunestage{} & field-permuted / 10 & \modelstage{} + \matchstage{} (permuted-prefix views) & fixed-order \tunestage{} & 2 + 2 epochs & robustness-oriented instantiation \\
Field-permuted \modelstage{} $\rightarrow$ field-permuted \tunestage{} & field-permuted / 10 & \modelstage{} & field-permuted \tunestage{} & matched steps & order-matched control \\
Field-permuted \modelstage{} + \matchstage{} $\rightarrow$ field-permuted \tunestage{} & field-permuted / 10 & \modelstage{} + \matchstage{} (permuted-prefix views) & field-permuted \tunestage{} & matched steps & order-matched recipe \\
Single-view field-permuted \modelstage{} & field-permuted / 1 & \modelstage{} & -- & standard \modelstage{} steps & pure order effect \\
Matched-update field-permuted \modelstage{} & field-permuted / 10 & \modelstage{} & -- & standard \modelstage{} steps & field-order permutation at fixed compute \\
Legacy-serialization \modelstage{} + \matchstage{} & legacy / 1 & \modelstage{} + \matchstage{} (disjoint views) & -- & 2 epochs & original serialization \\
\modelstage{} + order-aware rank & fixed-order / 1 & \modelstage{} + order-aware rank & -- & 2 epochs & order-aware alternative \\
\bottomrule
\end{tabular}}
\end{table}

The complete sweep also contains \tunestage{} descendants of the consistency-only and order-aware arms.
At 1.5B, \matchstage{} is directionally positive in the four model-and-match comparisons,
but only the \modelstage{}+\matchstage{}$\rightarrow$\tunestage{} comparison against
\modelstage{}$\rightarrow$\tunestage{} exceeds the embedding duplicate band; the aligned
larger-scale aggregate effects remain small. Comparing the two field-permuted Stage-1 parents
under standard versus field-permuted \tunestage{} isolates the supervised presentation; at 1.5B,
the standard presentation improves \textsc{Emb-8} by 0.012/0.014
(Tab.~\ref{tab:aggregate-ablation-controls}).

\subsection{Training Details}
\label{app:training-details}
Every standard Stage-1 arm uses two
data epochs, effective batch 16, AdamW ($\beta_1=0.9,\beta_2=0.95$, weight decay 0.1), learning
rate $10^{-5}$, 100 warmup steps, cosine decay, maximum length 512, bfloat16, and seed 42. Larger
models use FSDP while preserving the same effective batch. Appx.~\ref{app:arms} contains the
complete arm registry.

The 512-token budget truncates some serialized records. On MiniBooNE, all 3,000 test rows contain
590--726 tokens (mean 708.5), so every row is truncated; the encoded prefix contains 35.8 of 50
complete fields on average, and 91.5\% of raw truncations end inside the next key-value unit.

We also audited the field-order permutation construction on the first 3,000 MiniBooNE training rows,
using ten deterministic pairs per row (30,000 pairs total). Every pair exposed different
complete-field sets; each view exposed 35.7 of 50 complete fields on average, the mean set Jaccard
similarity was 0.556, and the mean symmetric difference between paired prefixes was 20.4 fields.
These measurements confirm that permuted-prefix views change both field order and visible content.

\section{Extended Overall Results}
\label{app:perdataset}

This section provides the complete controls and per-dataset metrics underlying the compact
main-paper comparisons.

\subsection{Matched recipe and ablation controls}

\begin{table}[H]
\caption{\textbf{Matched complete-record comparison for the final \matchstage{} intervention.}
Both recipes use field-permuted \modelstage{} followed by the same \tunestage{} interface; the
second additionally uses \matchstage{}. The near-identical mean motivates evaluating Match through
its partial-view robustness role rather than complete-record aggregate accuracy. Higher is better.}
\label{tab:m24-m26}
\centering
\scriptsize
\resizebox{\textwidth}{!}{%
\begin{tabular}{lrrrrrrrrr}
\toprule
Recipe & MiniBooNE & HIGGS & Supercond. & YearPred & Naval & CPU act. & Protein & USDA & Mean 8 \\
\midrule
Field-permuted \modelstage{} $\rightarrow$ \tunestage{} &
.963 & \textbf{.697} & \textbf{.817} & \textbf{.224} & .960 & \textbf{.970} & .424 & \textbf{.969} & \textbf{.753} \\
Field-permuted \modelstage{} + \matchstage{} $\rightarrow$ \tunestage{} &
\textbf{.968} & .691 & .806 & .219 & \textbf{.972} & .965 & \textbf{.434} & .967 & .753 \\
\bottomrule
\end{tabular}}
\end{table}

\begin{table}[H]
\caption{\textbf{Aggregate controls used in the ablation discussion.}
\textsc{Emb-8} is the eight-dataset frozen-probe aggregate; imputation is the corresponding
eight-dataset within-10\% accuracy. The final two rows use field-permuted records during
\tunestage{} to isolate the supervised interface. Higher is better.}
\label{tab:aggregate-ablation-controls}
\centering
\small
\resizebox{\textwidth}{!}{%
\begin{tabular}{lrr}
\toprule
Recipe & \textsc{Emb-8} $\uparrow$ & Imputation $\uparrow$ \\
\midrule
Frozen decoder & .688 & .106 \\
\modelstage{} & .730 & .401 \\
\matchstage{} only & .614 & .033 \\
\modelstage{}+\matchstage{} & .733 & .401 \\
Field-permuted \modelstage{} & .739 & .448 \\
Field-permuted \modelstage{}+\matchstage{} & .740 & .445 \\
\midrule
\modelstage{} $\rightarrow$ \tunestage{} & .736 & .145 \\
\modelstage{}+\matchstage{} $\rightarrow$ \tunestage{} & .743 & .145 \\
Field-permuted \modelstage{} $\rightarrow$ \tunestage{} & .753 & .221 \\
Field-permuted \modelstage{}+\matchstage{} $\rightarrow$ \tunestage{} & .753 & .234 \\
Field-permuted \modelstage{} $\rightarrow$ field-permuted \tunestage{} & .741 & -- \\
Field-permuted \modelstage{}+\matchstage{} $\rightarrow$ field-permuted \tunestage{} & .740 & -- \\
\bottomrule
\end{tabular}}
\end{table}

\subsubsection{Sensitivity to the Loss Weights}
\label{app:loss-weight-sensitivity}
\label{sec:loss-weights}

Eq.~\eqref{eq:adapt} fixes one pair of coefficients. Writing the \modelstage{}+\matchstage{}
objective in general form as
\begin{equation}
    \mathcal{L}_{\mathrm{Model+Match}}=\lambda_1\,\recordloss + \lambda_2\,\consloss,
    \label{eq:adapt-general}
\end{equation}
the reported recipe is $\lambda_1=0.5$, $\lambda_2=1.0$. Because Adam normalizes each parameter's
update by its own gradient scale, multiplying both coefficients by a constant leaves the update
nearly unchanged, so the operative hyperparameter is the ratio $\lambda_2/\lambda_1$. We therefore
sweep that ratio at fixed $\lambda_1=0.5$ and include one row at doubled scale to test the
reduction.

\begin{table}[h]
\caption{\textbf{The reported consistency weight is at the optimum on the dataset where
\matchstage{} has an effect, and irrelevant on one where it does not.} Primary embedding metric
($R^2$) after \modelstage{}+\matchstage{} at 1.5B with permuted-prefix views, varying only the loss
coefficients of Eq.~\eqref{eq:adapt-general}. The reported setting is in bold. The last two rows
are controls: $\lambda_1{=}0$ removes the record term, and $\lambda_1{=}1.0,\lambda_2{=}2.0$
preserves the ratio at doubled scale. The same-configuration measurement
spread on this metric is $0.005$, so differences below that are not resolved.}
\label{tab:loss-weights}
\centering
\small
\begin{tabular}{rrcc}
\toprule
$\lambda_1$ & $\lambda_2$ & Protein $R^2 \uparrow$ & USDA $R^2 \uparrow$ \\
\midrule
0.5 & 0.00 & .396 & .942 \\
0.5 & 0.25 & .411 & .940 \\
0.5 & 0.50 & .415 & .939 \\
\best{0.5} & \best{1.00} & \best{.420} & .937 \\
0.5 & 2.00 & .416 & .938 \\
0.5 & 4.00 & .414 & .938 \\
\midrule
0.0 & 1.00 & .248 & .878 \\
1.0 & 2.00 & .420 & .938 \\
\bottomrule
\end{tabular}
\end{table}

On Protein, where \matchstage{} changes the embedding metric, the sweep is single-peaked at the
reported coefficients: $R^2$ rises from $.396$ at $\lambda_2{=}0$ to $.420$ at $\lambda_2{=}1.0$,
then falls to $.414$ at $\lambda_2{=}4.0$. Halving or quadrupling $\lambda_2$ costs $.005$ and
$.006$ respectively, while removing the term costs $.024$; the choice is therefore neither
arbitrary nor delicate, since any $\lambda_2 \in [0.5, 2.0]$ stays within $.005$ of the best value.
On USDA every setting lies within measurement spread of every other, including the removal of
\matchstage{} altogether. Read together, the two columns indicate that $\lambda_2$ is worth tuning
only where the consistency term is active at all, and they explain why \matchstage{} looks inert in
aggregate: averaging a dataset where it is worth $+.024$ with several where it is worth nothing
yields a small mean. Held-out record perplexity is unchanged across the swept rows ($2.13$--$2.14$
on Protein), so the gain is not accompanied by worse held-out record likelihood. The two control rows behave as
expected: dropping $\recordloss$ collapses the representation ($R^2$ $.248$, perplexity $92.6$),
confirming that no weighting of $\consloss$ substitutes for the record term, and doubling both
coefficients reproduces the reported row to within $.001$, supporting the reduction to a single
ratio. This sweep covers two regression datasets at one scale and one seed, so it bounds the
sensitivity of the reported setting rather than establishing a globally optimal weight.

\subsection{Complete per-dataset target-prediction results}
Tabs.~\ref{tab:embedding-detail-cls}--\ref{tab:embedding-detail-pearson} report the complete
per-dataset results. \textit{Resources} gives parameter count for neural models, the input resource
for non-neural baselines, and API access for Cohere. The frozen-decoder recipe and
Qwen2.5-1.5B are the same checkpoint. The Cohere probe uses 10,000 training records; all other
zero-shot probes use 24,000.

\begin{table}[H]
\caption{Detailed classification-probe results. ACC and ROC--AUC are reported for MiniBooNE and HIGGS; higher is better. Mean is the arithmetic mean of the four displayed metrics. \recipedefs}
\label{tab:embedding-detail-cls}
\centering
\scriptsize
\setlength{\tabcolsep}{2.5pt}
\renewcommand{\arraystretch}{0.95}
\resizebox{\textwidth}{!}{%
\begin{tabular}{llrrrrr}
\toprule
\multirow{2}{*}{Model / recipe} & \multirow{2}{*}{Resources} & \multicolumn{2}{c}{MiniBooNE} & \multicolumn{2}{c}{HIGGS} & \multirow{2}{*}{Mean $\uparrow$} \\
\cmidrule(lr){3-4}\cmidrule(lr){5-6}
& & ACC $\uparrow$ & ROC--AUC $\uparrow$ & ACC $\uparrow$ & ROC--AUC $\uparrow$ & \\
\midrule
\multicolumn{7}{l}{\textit{floors / ceilings}} \\
\textbf{F2:} char-TF-IDF probe floor & 50k sparse features & 0.833 & 0.890 & 0.554 & 0.552 & .707 \\
\textbf{C1:} XGBoost raw-feature ceiling & raw features & 0.940 & 0.983 & 0.706 & 0.788 & .854 \\
\midrule
\multicolumn{7}{l}{\textit{zero-shot decoder LLMs (18 models)}} \\
Qwen2.5-1.5B & 1.5B & 0.898 & 0.956 & 0.617 & 0.655 & .781 \\
Qwen3-1.7B & 1.7B & 0.896 & 0.957 & 0.609 & 0.651 & .778 \\
Qwen3-4B & 4B & 0.904 & 0.960 & 0.607 & 0.642 & .778 \\
Qwen2.5-7B & 7B & 0.900 & 0.960 & 0.604 & 0.644 & .777 \\
Qwen3-8B & 8B & 0.897 & 0.955 & 0.609 & 0.636 & .774 \\
Qwen3-14B & 14B & 0.901 & 0.959 & 0.621 & 0.660 & .785 \\
Mistral-7B-v0.3 & 7B & 0.905 & 0.958 & 0.612 & 0.647 & .780 \\
OLMo-2-7B & 7B & 0.898 & 0.961 & 0.667 & 0.725 & .813 \\
phi-4 (14B) & 14B & 0.901 & 0.958 & 0.662 & 0.722 & .811 \\
DeepSeek-V2-Lite (16B MoE) & 16B MoE & 0.892 & 0.958 & 0.606 & 0.646 & .775 \\
gpt-oss-20b & 20B MoE & 0.891 & 0.954 & 0.618 & 0.668 & .783 \\
SmolLM3-3B & 3B & 0.901 & 0.959 & 0.646 & 0.705 & .803 \\
Falcon3-7B-Base & 7B & 0.900 & 0.959 & 0.566 & 0.607 & .758 \\
granite-3.3-8b-base & 8B & 0.901 & 0.961 & 0.615 & 0.655 & .783 \\
Mistral-Nemo-Base-12B & 12B & 0.896 & 0.956 & 0.610 & 0.662 & .781 \\
gemma-4-12B & 12B & 0.896 & 0.957 & 0.601 & 0.639 & .773 \\
Llama-3.1-8B & 8B & 0.900 & 0.960 & 0.657 & 0.712 & .807 \\
Qwen3.5-9B & 9.6B & 0.892 & 0.954 & 0.601 & 0.644 & .773 \\
\midrule
\multicolumn{7}{l}{\textit{purpose-built embedders (10 models)}} \\
bge-small-en-v1.5 & 33M & 0.838 & 0.905 & 0.560 & 0.580 & .721 \\
all-MiniLM-L6-v2 & 22M & 0.812 & 0.870 & 0.577 & 0.596 & .714 \\
gte-Qwen2-1.5B-instruct & 1.5B & 0.857 & 0.913 & 0.558 & 0.569 & .724 \\
e5-mistral-7b-instruct & 7B & 0.850 & 0.906 & 0.556 & 0.571 & .721 \\
Qwen3-Embedding-0.6B & 0.6B & 0.881 & 0.943 & 0.575 & 0.598 & .749 \\
Qwen3-Embedding-4B & 4B & 0.884 & 0.943 & 0.593 & 0.620 & .760 \\
Qwen3-Embedding-8B & 8B & 0.885 & 0.944 & 0.585 & 0.620 & .758 \\
gte-Qwen2-7B-instruct & 7B & 0.865 & 0.923 & 0.564 & 0.588 & .735 \\
NV-Embed-v2 & 7B & 0.891 & 0.952 & 0.576 & 0.608 & .757 \\
LLM2Vec-Mistral-7B & 7B & 0.892 & 0.955 & 0.588 & 0.619 & .763 \\
\midrule
\multicolumn{7}{l}{\textit{commercial embedding API (selected)}} \\
cohere.embed-v4 (Bedrock) & API & 0.868 & 0.930 & 0.616 & 0.664 & .770 \\
\midrule
\multicolumn{7}{l}{\textit{trained 1.5B recipes}} \\
Frozen decoder & 1.5B & 0.898 & 0.956 & 0.617 & 0.655 & .781 \\
\modelstage{} & 1.5B & 0.911 & 0.968 & 0.628 & 0.675 & .795 \\
\tunestage{} only & 1.5B & 0.907 & 0.963 & 0.629 & 0.682 & .795 \\
\modelstage{} $\rightarrow$ \tunestage{} & 1.5B & 0.917 & 0.969 & 0.633 & 0.691 & .802 \\
\modelstage{} + \matchstage{} $\rightarrow$ \tunestage{} & 1.5B & 0.913 & 0.969 & 0.636 & 0.695 & .803 \\
Field-permuted \modelstage{} + \matchstage{} $\rightarrow$ \tunestage{} & 1.5B & 0.913 & 0.968 & 0.630 & 0.691 & .801 \\
\bottomrule
\end{tabular}}
\end{table}

\begin{table}[H]
\caption{Detailed regression-probe $R^2$ by dataset; higher is better. Mean is the arithmetic mean over the six displayed datasets. \recipedefs}
\label{tab:embedding-detail-r2}
\centering
\scriptsize
\setlength{\tabcolsep}{2.5pt}
\renewcommand{\arraystretch}{0.95}
\resizebox{\textwidth}{!}{%
\begin{tabular}{llrrrrrrr}
\toprule
Model / recipe & Resources & Supercond. & USDA & CPU & Naval & Protein & YearPred & Mean $\uparrow$ \\
\midrule
\multicolumn{9}{l}{\textit{floors / ceilings}} \\
\textbf{F2:} char-TF-IDF probe floor & 50k sparse features & 0.865 & 0.860 & 0.281 & 0.904 & 0.201 & 0.005 & .519 \\
\textbf{C1:} XGBoost raw-feature ceiling & raw features & 0.916 & 0.997 & 0.985 & 0.996 & 0.623 & 0.277 & .799 \\
\midrule
\multicolumn{9}{l}{\textit{zero-shot decoder LLMs (18 models)}} \\
Qwen2.5-1.5B & 1.5B & 0.769 & 0.910 & 0.891 & 0.860 & 0.311 & 0.152 & .649 \\
Qwen3-1.7B & 1.7B & 0.778 & 0.920 & 0.886 & 0.885 & 0.316 & 0.155 & .657 \\
Qwen3-4B & 4B & 0.783 & 0.940 & 0.926 & 0.893 & 0.332 & 0.160 & .672 \\
Qwen2.5-7B & 7B & 0.814 & 0.939 & 0.945 & 0.927 & 0.323 & 0.158 & .684 \\
Qwen3-8B & 8B & 0.797 & 0.946 & 0.931 & 0.917 & 0.330 & 0.172 & .682 \\
Qwen3-14B & 14B & 0.809 & 0.951 & 0.947 & 0.929 & 0.342 & 0.156 & .689 \\
Mistral-7B-v0.3 & 7B & 0.813 & 0.950 & 0.948 & 0.929 & 0.355 & 0.172 & .695 \\
OLMo-2-7B & 7B & 0.817 & 0.941 & 0.921 & 0.930 & 0.329 & 0.160 & .683 \\
phi-4 (14B) & 14B & 0.810 & 0.954 & 0.923 & 0.933 & 0.340 & 0.143 & .684 \\
DeepSeek-V2-Lite (16B MoE) & 16B MoE & 0.778 & 0.926 & 0.921 & 0.837 & 0.334 & 0.167 & .661 \\
gpt-oss-20b & 20B MoE & 0.753 & 0.915 & 0.803 & 0.830 & 0.284 & 0.153 & .623 \\
SmolLM3-3B & 3B & 0.780 & 0.913 & 0.876 & 0.876 & 0.277 & 0.159 & .647 \\
Falcon3-7B-Base & 7B & 0.799 & 0.930 & 0.921 & 0.924 & 0.350 & 0.158 & .680 \\
granite-3.3-8b-base & 8B & 0.805 & 0.935 & 0.945 & 0.901 & 0.351 & 0.155 & .682 \\
Mistral-Nemo-Base-12B & 12B & 0.810 & 0.945 & 0.944 & 0.944 & 0.348 & 0.152 & .691 \\
gemma-4-12B & 12B & 0.813 & 0.952 & 0.922 & 0.921 & 0.353 & 0.144 & .684 \\
Llama-3.1-8B & 8B & 0.804 & 0.946 & 0.911 & 0.915 & 0.321 & 0.174 & .679 \\
Qwen3.5-9B & 9.6B & 0.807 & 0.939 & 0.928 & 0.897 & 0.349 & 0.149 & .678 \\
\midrule
\multicolumn{9}{l}{\textit{purpose-built embedders (10 models)}} \\
bge-small-en-v1.5 & 33M & 0.562 & 0.673 & 0.611 & 0.608 & 0.088 & 0.102 & .441 \\
all-MiniLM-L6-v2 & 22M & 0.623 & 0.664 & 0.683 & 0.624 & 0.110 & 0.026 & .455 \\
gte-Qwen2-1.5B-instruct & 1.5B & 0.574 & 0.670 & 0.954 & 0.745 & 0.258 & 0.014 & .536 \\
e5-mistral-7b-instruct & 7B & 0.662 & 0.662 & 0.942 & 0.810 & 0.310 & 0.013 & .567 \\
Qwen3-Embedding-0.6B & 0.6B & 0.668 & 0.781 & 0.755 & 0.743 & 0.195 & 0.117 & .543 \\
Qwen3-Embedding-4B & 4B & 0.690 & 0.868 & 0.868 & 0.716 & 0.257 & 0.110 & .585 \\
Qwen3-Embedding-8B & 8B & 0.770 & 0.903 & 0.893 & 0.849 & 0.275 & 0.118 & .635 \\
gte-Qwen2-7B-instruct & 7B & 0.642 & 0.692 & 0.945 & 0.795 & 0.325 & 0.017 & .569 \\
NV-Embed-v2 & 7B & 0.803 & 0.916 & 0.932 & 0.922 & 0.323 & 0.139 & .673 \\
LLM2Vec-Mistral-7B & 7B & 0.812 & 0.904 & 0.902 & 0.944 & 0.318 & 0.162 & .674 \\
\midrule
\multicolumn{9}{l}{\textit{commercial embedding API (selected)}} \\
cohere.embed-v4 (Bedrock) & API & 0.770 & 0.863 & 0.764 & 0.871 & 0.200 & 0.126 & .599 \\
\midrule
\multicolumn{9}{l}{\textit{trained 1.5B recipes}} \\
Frozen decoder & 1.5B & 0.769 & 0.910 & 0.891 & 0.860 & 0.311 & 0.152 & .649 \\
\modelstage{} & 1.5B & 0.808 & 0.914 & 0.954 & 0.920 & 0.390 & 0.213 & .700 \\
\tunestage{} only & 1.5B & 0.773 & 0.939 & 0.904 & 0.884 & 0.380 & 0.185 & .678 \\
\modelstage{} $\rightarrow$ \tunestage{} & 1.5B & 0.804 & 0.943 & 0.934 & 0.946 & 0.401 & 0.198 & .704 \\
\modelstage{} + \matchstage{} $\rightarrow$ \tunestage{} & 1.5B & 0.797 & 0.950 & 0.949 & 0.959 & 0.413 & 0.211 & .713 \\
Field-permuted \modelstage{} + \matchstage{} $\rightarrow$ \tunestage{} & 1.5B & 0.806 & 0.967 & 0.965 & 0.972 & 0.434 & 0.219 & .727 \\
\bottomrule
\end{tabular}}
\end{table}

\begin{table}[H]
\caption{Detailed regression-probe MAE by dataset; lower is better. Naval MAE is multiplied by $10^4$; other columns retain their target units. Mean is the arithmetic mean of these displayed values and is a descriptive aggregate because units differ across datasets. \recipedefs}
\label{tab:embedding-detail-mae}
\centering
\scriptsize
\setlength{\tabcolsep}{2.5pt}
\renewcommand{\arraystretch}{0.95}
\resizebox{\textwidth}{!}{%
\begin{tabular}{llrrrrrrr}
\toprule
Model / recipe & Resources & Supercond. & USDA & CPU & Naval $\times 10^4$ & Protein & YearPred & Mean $\downarrow$ \\
\midrule
\multicolumn{9}{l}{\textit{floors / ceilings}} \\
\textbf{F2:} char-TF-IDF probe floor & 50k sparse features & 8.60 & 44.30 & 8.83 & 26.03 & 4.46 & 8.13 & 16.725 \\
\textbf{C1:} XGBoost raw-feature ceiling & raw features & 5.59 & 4.75 & 1.48 & 6.57 & 2.70 & 6.62 & 4.618 \\
\midrule
\multicolumn{9}{l}{\textit{zero-shot decoder LLMs (18 models)}} \\
Qwen2.5-1.5B & 1.5B & 12.10 & 35.93 & 4.29 & 41.12 & 4.11 & 7.33 & 17.480 \\
Qwen3-1.7B & 1.7B & 11.84 & 34.51 & 4.45 & 37.67 & 4.10 & 7.32 & 16.648 \\
Qwen3-4B & 4B & 11.63 & 29.29 & 3.60 & 35.96 & 4.08 & 7.27 & 15.305 \\
Qwen2.5-7B & 7B & 10.58 & 28.97 & 3.09 & 28.30 & 4.10 & 7.30 & 13.723 \\
Qwen3-8B & 8B & 11.17 & 27.06 & 3.50 & 31.27 & 4.08 & 7.30 & 14.063 \\
Qwen3-14B & 14B & 10.56 & 25.84 & 3.06 & 28.53 & 4.02 & 7.36 & 13.228 \\
Mistral-7B-v0.3 & 7B & 10.61 & 26.14 & 3.00 & 28.40 & 3.98 & 7.26 & 13.232 \\
OLMo-2-7B & 7B & 10.36 & 28.95 & 3.75 & 28.93 & 4.05 & 7.28 & 13.887 \\
phi-4 (14B) & 14B & 10.57 & 25.87 & 3.62 & 27.26 & 4.04 & 7.42 & 13.130 \\
DeepSeek-V2-Lite (16B MoE) & 16B MoE & 11.97 & 33.29 & 3.68 & 44.94 & 4.01 & 7.22 & 17.518 \\
gpt-oss-20b & 20B MoE & 12.51 & 37.31 & 5.93 & 44.57 & 4.23 & 7.36 & 18.652 \\
SmolLM3-3B & 3B & 11.76 & 35.98 & 4.63 & 38.28 & 4.23 & 7.21 & 17.015 \\
Falcon3-7B-Base & 7B & 11.08 & 32.48 & 3.66 & 30.08 & 3.98 & 7.31 & 14.765 \\
granite-3.3-8b-base & 8B & 10.90 & 29.12 & 3.02 & 34.57 & 3.99 & 7.33 & 14.822 \\
Mistral-Nemo-Base-12B & 12B & 10.56 & 27.65 & 3.11 & 26.12 & 3.99 & 7.29 & 13.120 \\
gemma-4-12B & 12B & 10.72 & 26.17 & 3.62 & 30.94 & 3.98 & 7.39 & 13.803 \\
Llama-3.1-8B & 8B & 10.97 & 26.96 & 3.80 & 32.27 & 4.05 & 7.20 & 14.208 \\
Qwen3.5-9B & 9.6B & 10.90 & 30.13 & 3.54 & 34.78 & 3.98 & 7.38 & 15.118 \\
\midrule
\multicolumn{9}{l}{\textit{purpose-built embedders (10 models)}} \\
bge-small-en-v1.5 & 33M & 17.66 & 71.96 & 7.94 & 71.93 & 5.03 & 7.51 & 30.338 \\
all-MiniLM-L6-v2 & 22M & 16.24 & 72.18 & 7.22 & 69.68 & 4.96 & 8.00 & 29.713 \\
gte-Qwen2-1.5B-instruct & 1.5B & 17.19 & 74.92 & 2.75 & 57.93 & 4.37 & 8.11 & 27.545 \\
e5-mistral-7b-instruct & 7B & 14.56 & 73.33 & 3.11 & 48.02 & 4.15 & 8.09 & 25.210 \\
Qwen3-Embedding-0.6B & 0.6B & 14.85 & 58.94 & 6.55 & 56.88 & 4.55 & 7.42 & 24.865 \\
Qwen3-Embedding-4B & 4B & 14.20 & 43.52 & 4.89 & 58.65 & 4.32 & 7.46 & 22.173 \\
Qwen3-Embedding-8B & 8B & 12.00 & 37.92 & 4.28 & 41.99 & 4.19 & 7.52 & 17.983 \\
gte-Qwen2-7B-instruct & 7B & 15.04 & 70.13 & 3.06 & 50.89 & 4.10 & 8.00 & 25.203 \\
NV-Embed-v2 & 7B & 10.90 & 34.69 & 3.45 & 28.50 & 4.07 & 7.38 & 14.832 \\
LLM2Vec-Mistral-7B & 7B & 10.75 & 36.29 & 4.11 & 25.30 & 4.08 & 7.27 & 14.633 \\
\midrule
\multicolumn{9}{l}{\textit{commercial embedding API (selected)}} \\
cohere.embed-v4 (Bedrock) & API & 12.12 & 46.63 & 6.43 & 38.17 & 4.55 & 7.41 & 19.218 \\
\midrule
\multicolumn{9}{l}{\textit{trained 1.5B recipes}} \\
Frozen decoder & 1.5B & 12.10 & 35.93 & 4.29 & 41.12 & 4.11 & 7.33 & 17.480 \\
\modelstage{} & 1.5B & 10.72 & 35.60 & 2.80 & 30.03 & 3.81 & 6.99 & 14.992 \\
\tunestage{} only & 1.5B & 11.91 & 29.31 & 4.14 & 35.04 & 3.81 & 7.09 & 15.217 \\
\modelstage{} $\rightarrow$ \tunestage{} & 1.5B & 10.94 & 29.08 & 3.42 & 25.01 & 3.76 & 7.00 & 13.202 \\
\modelstage{} + \matchstage{} $\rightarrow$ \tunestage{} & 1.5B & 11.11 & 26.51 & 2.94 & 20.86 & 3.68 & 6.93 & 12.005 \\
Field-permuted \modelstage{} + \matchstage{} $\rightarrow$ \tunestage{} & 1.5B & 10.51 & 20.42 & 2.43 & 18.05 & 3.64 & 6.96 & 10.335 \\
\bottomrule
\end{tabular}}
\end{table}

\begin{table}[H]
\caption{Detailed regression-probe Pearson $r$ by dataset; higher is better. Mean is the arithmetic mean over the six displayed datasets. \recipedefs}
\label{tab:embedding-detail-pearson}
\centering
\scriptsize
\setlength{\tabcolsep}{2.5pt}
\renewcommand{\arraystretch}{0.95}
\resizebox{\textwidth}{!}{%
\begin{tabular}{llrrrrrrr}
\toprule
Model / recipe & Resources & Supercond. & USDA & CPU & Naval & Protein & YearPred & Mean $\uparrow$ \\
\midrule
\multicolumn{9}{l}{\textit{floors / ceilings}} \\
\textbf{F2:} char-TF-IDF probe floor & 50k sparse features & 0.930 & 0.928 & 0.538 & 0.951 & 0.459 & 0.081 & .648 \\
\textbf{C1:} XGBoost raw-feature ceiling & raw features & 0.957 & 0.999 & 0.993 & 0.998 & 0.790 & 0.527 & .877 \\
\midrule
\multicolumn{9}{l}{\textit{zero-shot decoder LLMs (18 models)}} \\
Qwen2.5-1.5B & 1.5B & 0.878 & 0.954 & 0.945 & 0.928 & 0.558 & 0.391 & .776 \\
Qwen3-1.7B & 1.7B & 0.883 & 0.959 & 0.942 & 0.941 & 0.564 & 0.395 & .781 \\
Qwen3-4B & 4B & 0.885 & 0.970 & 0.962 & 0.945 & 0.577 & 0.403 & .790 \\
Qwen2.5-7B & 7B & 0.903 & 0.969 & 0.972 & 0.963 & 0.569 & 0.401 & .796 \\
Qwen3-8B & 8B & 0.893 & 0.972 & 0.965 & 0.958 & 0.575 & 0.419 & .797 \\
Qwen3-14B & 14B & 0.901 & 0.975 & 0.973 & 0.964 & 0.586 & 0.403 & .800 \\
Mistral-7B-v0.3 & 7B & 0.902 & 0.975 & 0.973 & 0.964 & 0.596 & 0.418 & .805 \\
OLMo-2-7B & 7B & 0.905 & 0.970 & 0.960 & 0.965 & 0.574 & 0.407 & .797 \\
phi-4 (14B) & 14B & 0.900 & 0.977 & 0.961 & 0.966 & 0.584 & 0.393 & .797 \\
DeepSeek-V2-Lite (16B MoE) & 16B MoE & 0.883 & 0.962 & 0.960 & 0.915 & 0.579 & 0.410 & .785 \\
gpt-oss-20b & 20B MoE & 0.868 & 0.957 & 0.897 & 0.911 & 0.534 & 0.399 & .761 \\
SmolLM3-3B & 3B & 0.884 & 0.956 & 0.937 & 0.936 & 0.528 & 0.402 & .774 \\
Falcon3-7B-Base & 7B & 0.895 & 0.965 & 0.960 & 0.961 & 0.593 & 0.401 & .796 \\
granite-3.3-8b-base & 8B & 0.898 & 0.967 & 0.972 & 0.950 & 0.593 & 0.399 & .797 \\
Mistral-Nemo-Base-12B & 12B & 0.901 & 0.972 & 0.972 & 0.972 & 0.591 & 0.395 & .801 \\
gemma-4-12B & 12B & 0.902 & 0.976 & 0.960 & 0.960 & 0.595 & 0.386 & .797 \\
Llama-3.1-8B & 8B & 0.897 & 0.973 & 0.955 & 0.957 & 0.569 & 0.419 & .795 \\
Qwen3.5-9B & 9.6B & 0.899 & 0.969 & 0.963 & 0.947 & 0.592 & 0.394 & .794 \\
\midrule
\multicolumn{9}{l}{\textit{purpose-built embedders (10 models)}} \\
bge-small-en-v1.5 & 33M & 0.750 & 0.821 & 0.785 & 0.780 & 0.300 & 0.322 & .626 \\
all-MiniLM-L6-v2 & 22M & 0.790 & 0.816 & 0.829 & 0.791 & 0.333 & 0.163 & .620 \\
gte-Qwen2-1.5B-instruct & 1.5B & 0.758 & 0.819 & 0.977 & 0.863 & 0.509 & 0.144 & .678 \\
e5-mistral-7b-instruct & 7B & 0.814 & 0.816 & 0.971 & 0.900 & 0.557 & 0.163 & .704 \\
Qwen3-Embedding-0.6B & 0.6B & 0.819 & 0.884 & 0.869 & 0.863 & 0.444 & 0.342 & .704 \\
Qwen3-Embedding-4B & 4B & 0.833 & 0.932 & 0.933 & 0.848 & 0.510 & 0.338 & .732 \\
Qwen3-Embedding-8B & 8B & 0.878 & 0.950 & 0.945 & 0.922 & 0.532 & 0.350 & .763 \\
gte-Qwen2-7B-instruct & 7B & 0.802 & 0.832 & 0.972 & 0.891 & 0.570 & 0.187 & .709 \\
NV-Embed-v2 & 7B & 0.897 & 0.957 & 0.965 & 0.960 & 0.568 & 0.383 & .788 \\
LLM2Vec-Mistral-7B & 7B & 0.901 & 0.951 & 0.950 & 0.972 & 0.564 & 0.408 & .791 \\
\midrule
\multicolumn{9}{l}{\textit{commercial embedding API (selected)}} \\
cohere.embed-v4 (Bedrock) & API & 0.878 & 0.929 & 0.875 & 0.933 & 0.448 & 0.360 & .737 \\
\midrule
\multicolumn{9}{l}{\textit{trained 1.5B recipes}} \\
Frozen decoder & 1.5B & 0.878 & 0.954 & 0.945 & 0.928 & 0.558 & 0.391 & .776 \\
\modelstage{} & 1.5B & 0.900 & 0.956 & 0.977 & 0.959 & 0.626 & 0.462 & .813 \\
\tunestage{} only & 1.5B & 0.880 & 0.969 & 0.951 & 0.940 & 0.617 & 0.431 & .798 \\
\modelstage{} $\rightarrow$ \tunestage{} & 1.5B & 0.897 & 0.971 & 0.967 & 0.973 & 0.633 & 0.445 & .814 \\
\modelstage{} + \matchstage{} $\rightarrow$ \tunestage{} & 1.5B & 0.894 & 0.975 & 0.974 & 0.979 & 0.643 & 0.460 & .821 \\
Field-permuted \modelstage{} + \matchstage{} $\rightarrow$ \tunestage{} & 1.5B & 0.899 & 0.983 & 0.983 & 0.986 & 0.659 & 0.468 & .830 \\
\bottomrule
\end{tabular}}
\end{table}

\subsection{Generality to records with substantial text and categorical fields}
\label{app:mixed-type}

The eight datasets in Tab.~\ref{tab:embedding-per-dataset} are numerical by construction: measured
over field values, $272/272$ fields are numeric, so the benchmark says nothing about records whose
fields are largely words, which is common in real tabular data. We therefore re-ran the complete
recipe block of Tab.~\ref{tab:embedding-per-dataset} on two additional datasets, holding the
recipe, the serialization, the batch size, and the two-epoch budget exactly as in the main
experiments and changing only the data.

\textbf{Census-KDD} (OpenML 4535; 30{,}000 records, 41 fields, 28 of them categorical) is the
text-heavy case: \textbf{95.3\%} of the characters in a serialized record are non-numeric, with
values such as \texttt{State government} and \texttt{Bachelors degree(BA AB BS)}. It is the direct
test of whether the method survives text. \textbf{NYC-Taxi} (OpenML 42729, tip amount; 30{,}000
records, 18 fields, \textbf{9 of them declared categorical}) contains also text value: \textbf{3.0\%} of its value characters are non-numeric, and its frozen
zero-shot probe reaches $R^2=0.433$, leaving room for a training signal to register. Following Tab.~\ref{tab:embedding-per-dataset}, the
classification dataset reports ROC--AUC and the regression dataset reports $R^2$, and
\textsc{Emb-2} is their unweighted mean.

\begin{table}[h]
\caption{The complete \methodname{} recipe block re-run on records that are not numerical, in the
format of Tab.~\ref{tab:embedding-per-dataset}. The percentage after each dataset is the share of
non-numeric characters in a serialized record, measured over field values. \textsc{Emb-2} is the
unweighted mean of the two columns. All trained recipes use a Qwen2.5-1.5B backbone, one seed, and
the identical configuration reported in the main results; bold marks the best value in each column.
On the text-heavy dataset every recipe stays at or above the frozen baseline and the best recipe
exceeds it, but the recipe ordering is compressed because a frozen decoder already reaches $0.941$
there. On the dataset with headroom the ordering of Tab.~\ref{tab:embedding-per-dataset} reappears
in full.}
\label{tab:mixed-type}
\centering
\scriptsize
\setlength{\tabcolsep}{3pt}
\begin{tabular}{lrrr}
\toprule
\multirow{2}{*}{Model / recipe}
& \multicolumn{1}{c}{Classification: ROC--AUC $\uparrow$}
& \multicolumn{1}{c}{Regression: $R^2 \uparrow$}
& \multirow{2}{*}{\textsc{Emb-2} $\uparrow$} \\
\cmidrule(lr){2-2}\cmidrule(lr){3-3}
& Census-KDD (95.3\% text) & NYC-Taxi (3.0\% text) & \\
\midrule
\multicolumn{4}{l}{\textit{Frozen decoder LMs (mean-pooled hidden states)}} \\
Qwen3-8B & .940 & .444 & .692 \\
Gemma-4-12B & .937 & .490 & .713 \\
\midrule
\rowcolor{blue!6}
\multicolumn{4}{l}{\textit{\methodname{} framework (Qwen2.5-1.5B backbone)}} \\
Zero-shot frozen decoder & .941 & .433 & .687 \\
\tunestage{} only & .942 & .437 & .690 \\
\rowcolor{blue!6}
\modelstage{} $\rightarrow$ \tunestage{} & .940 & .485 & .712 \\
\rowcolor{blue!6}
\modelstage{} + \matchstage{} $\rightarrow$ \tunestage{} & .936 & .497 & .717 \\
\rowcolor{blue!6}
Field-permuted \modelstage{} $\rightarrow$ \tunestage{} & \textbf{.946} & .522 & .734 \\
\rowcolor{blue!6}
Field-permuted \modelstage{} + \matchstage{} $\rightarrow$ \tunestage{} & .945 & \textbf{.530} & \textbf{.738} \\
\bottomrule
\end{tabular}
\end{table}

\myparagraph{The method transfers to text-heavy records; resolving its components needs headroom.}
On Census-KDD, where more than nine in ten characters of a record are words, no recipe degrades the
representation: every two-stage recipe lands within $0.006$ ROC--AUC of the frozen $0.941$ baseline,
and the best recipe, field-permuted \modelstage{}$\rightarrow$\tunestage{}, reaches $0.946$ --
above the frozen 1.5B baseline and above both the frozen Qwen3-8B ($0.940$) and Gemma-4-12B
($0.937$) references. The gain is small in absolute terms ($+0.005$, just past the $0.005$
same-configuration spread reported for Tab.~\ref{tab:loss-weights}), and the differences among
recipes are not resolvable: a frozen decoder already scores $0.941$ on this task, so little of the
metric is available to move. That ceiling is a property of the task rather than of the field types,
and it appears in the main results too -- MiniBooNE moves from $0.956$ to $0.969$ while Protein, far
from saturation, moves from $0.311$ to $0.434$.

NYC-Taxi isolates that confound. Its records are almost entirely numeric, but its frozen probe is
far from the ceiling, and there the ordering of Tab.~\ref{tab:embedding-per-dataset} reappears
intact: \modelstage{}$\rightarrow$\tunestage{} adds $0.052\ R^2$ over zero-shot while \tunestage{}
alone adds $0.004$, field-order permutation adds a further $0.037$, and \matchstage{} adds $0.012$
at fixed order and $0.008$ under permutation -- all above the $0.005$ noise floor, with the
field-permuted \matchstage{} variant best at $0.530$, the same row that is best in
Tab.~\ref{tab:embedding-per-dataset}. Read together, the two datasets separate two questions that
the original benchmark conflates: whether the method tolerates non-numeric fields, which these
results support, and whether its individual components can be distinguished on a given task, which
requires that the task not already be solved by a frozen decoder. We did not observe a case where
text or categorical fields drove a recipe below the frozen baseline. The scope is two datasets, one
backbone, and one seed, so these results bound the concern rather than settle the size of the
effect on mixed-type data.

\section{Component Evidence and Mechanism Diagnostics}
\label{app:mechanism}

This section groups the diagnostic evidence by the capability associated with each training component. The complete protocols follow the corresponding interpretation so that each claim, measurement, and boundary can be read together. The digit-sensitivity and incomplete-view analyses are scoped diagnostics; their dataset and sample scopes are stated with each result.

\myparagraph{Paired record-bootstrap protocol.} For the digit-sensitivity and field-dropout diagnostics, we use 10,000 paired percentile-bootstrap resamples generated with NumPy PCG64 seed 20260925. The record is the sampling unit, and each resample uses the same indices across compared recipes, perturbations, and views. ROC--AUC is recomputed within every resample while the fitted probe remains fixed. The resulting intervals quantify evaluation-record uncertainty conditional on the trained checkpoints, not variation across training seeds.

\subsection{\modelstage{}: numerical sensitivity and cross-field context}

\modelstage{} provides the broadest Stage-1 gains. It raises \textsc{Emb-8} from 0.688 to 0.730
and masked-value completion from 0.106 to 0.401
(Tabs.~\ref{tab:embedding-per-dataset} and~\ref{tab:imputation-completion-1p5b}).
It also increases the correct-context gain from 0.058 to 0.167, while field-permuted
\modelstage{} reaches 0.271 (Tab.~\ref{tab:cross-field-context}). These results show that the
decoder becomes better at completing numerical values and relies more on the surrounding
measurements from the same example.

The digit perturbation diagnostic asks whether representations respond more strongly to changes
with larger numerical consequences. The leading-to-trailing sensitivity ratio rises from
$2.08\times$ for the frozen checkpoint to $13.83\times$ after fixed-order \modelstage{} and
$17.82\times$ after field-permuted \modelstage{}. The paired Model-minus-frozen difference is 11.75 (95\% CI [7.04, 17.09]); the additional field-permutation difference is 3.99 ([$-3.91$, 13.17]) and is unresolved. This is a scoped sensitivity proxy rather than a claim of faithful magnitude geometry: it is evaluated on MiniBooNE, and the absolute cosine distances are small.

\begin{figure}[H]
    \centering
    \includegraphics[width=\textwidth]{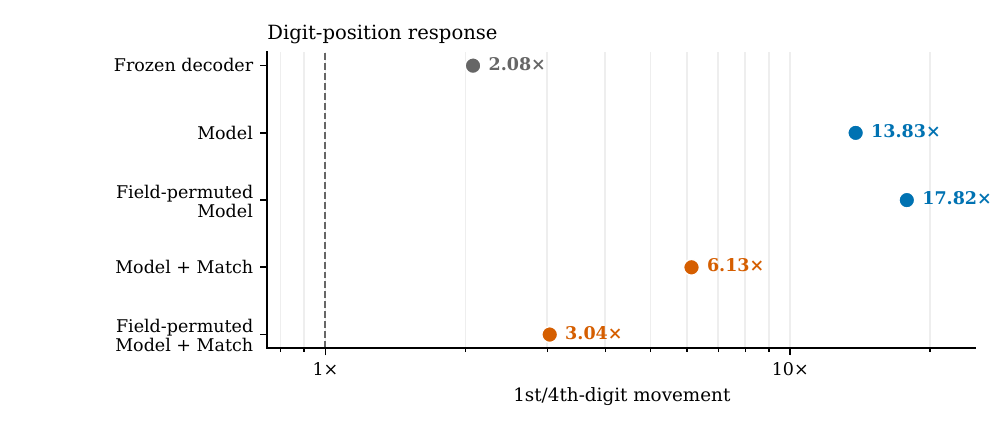}
    \caption{\textbf{Digit-position sensitivity across selected 1.5B arms.}
    \modelstage{} produces the largest increase over the frozen checkpoint: both \modelstage{}
    and field-permuted \modelstage{} become more sensitive to a first-digit edit. Adding
    \matchstage{} contracts the proxy for both \modelstage{} and field-permuted \modelstage{}.
    Every edited unit is complete in the encoded input and every variant changes one tokenizer token.
    MiniBooNE, 200 records; 95\% paired record-bootstrap intervals are reported in Tab.~\ref{tab:digit-sensitivity}. \recipedefs}
    \label{fig:digit-sensitivity}
\end{figure}

\subsubsection{Digit-position sensitivity protocol and full results}

For each of 200 MiniBooNE test records, we select a numeric field only after tokenizing the
fixed-order serialization. The selected key-value unit must be complete inside the 512-token input
for the original and all four variants. We create four copies by changing one of its first four
significant digits by one while keeping the serialized character length fixed; every variant must
change the encoded token sequence, and at least one changed token must overlap the selected unit.
All retained perturbations change exactly one token with the Qwen2.5 tokenizer. Let $d_p$ be the
cosine distance between the original embedding and the copy edited at
position $p$. We report $\bar d_1/\bar d_4$, the ratio of mean distances across records, and the
Spearman correlation between log absolute value change and embedding distance pooled over four
dependent perturbations per record. The same records and perturbations are used for every
checkpoint, with tokenizer identity checked against the reference token IDs. Because the absolute
distances are small and the ratio can be sensitive to its
denominator, we treat this as an exploratory diagnostic and do not interpret the distance scale.
The raw case-level artifact, including token spans, IDs, and model hashes, is
\texttt{learnings/digit\_sensitivity\_corrected\_20260913.json}.

\begin{table}[H]
\caption{Complete MiniBooNE digit-position diagnostic. The input first-versus-fourth value-change
ratio is approximately 1000$\times$; it is a reference contrast, not an expected embedding ratio.
Larger ratios indicate greater leading-digit sensitivity under this proxy. Intervals use 10,000
paired percentile-bootstrap resamples over records and are conditional on the evaluated checkpoints.}
\label{tab:digit-sensitivity}
\centering
\small
\begin{tabular}{lrrr}
\toprule
Recipe & $\bar d_1/\bar d_4$ $\uparrow$ & 95\% CI & Spearman $\rho$ $\uparrow$ \\
\midrule
Frozen decoder & 2.078 & [1.135, 3.761] & 0.261 \\
\modelstage{} & 13.832 & [9.363, 19.260] & 0.342 \\
\matchstage{} only & 1.008 & [0.871, 1.162] & -0.011 \\
\modelstage{} + \matchstage{} & 6.134 & [4.472, 8.106] & 0.333 \\
Field-permuted \modelstage{} & 17.822 & [9.559, 28.420] & 0.394 \\
Field-permuted \modelstage{} + \matchstage{} & 3.038 & [1.266, 8.038] & 0.376 \\
\modelstage{} $\rightarrow$ \tunestage{} & 4.434 & [3.217, 5.803] & 0.413 \\
\modelstage{} + \matchstage{} $\rightarrow$ \tunestage{} & 3.698 & [2.127, 5.960] & 0.389 \\
\midrule
Underlying value change & 1000.00 & -- & -- \\
\bottomrule
\end{tabular}
\end{table}

\subsection{\matchstage{}: stability and incomplete-record prediction}

\matchstage{} consistently contracts paired views, but whether that stability remains predictive
depends on the view construction. With permuted-prefix training, adding \matchstage{} reduces
full-to-incomplete displacement from 0.00468 to 0.000306 and raises mean incomplete-view AUC from
0.7192 to 0.7584. The paired displacement difference is $-0.004372$ (95\% CI [$-0.004390$, $-0.004355$]), equivalent to a $15.305\times$ reduction (95\% CI [$15.257\times$, $15.354\times$]); the paired incomplete-view AUC difference is $+0.0392$ ([0.0233, 0.0555]). The full-view AUC difference is unresolved at $+0.00145$ ([$-0.00453$, 0.00763]). With disjoint-half views, displacement also falls, but incomplete-view AUC drops
from 0.5841 to 0.5331, a paired difference of $-0.0510$ ([$-0.0822$, $-0.0200$]). A plausible explanation is that permuted-prefix views retain overlapping
conditional information, whereas disjoint halves can contain incompatible predictive fields.
The current experiment supports this view-compatibility hypothesis but does not establish a
general rule for constructing view pairs.

\matchstage{} should therefore be interpreted as a conditional robustness regularizer, not a general complete-record accuracy stage. Accordingly, \matchstage{} is not required by the core \methodname{} path; it is an optional extension for settings in which robustness across partial record views is a deployment requirement. The matched field-permuted recipes reach the same
\textsc{Emb-8} after \tunestage{} (Tab.~\ref{tab:m24-m26}), while Stage-1 completion changes only from 0.448 to 0.445 (Tab.~\ref{tab:imputation-completion-1p5b}).

\begin{figure}[H]
    \centering
    \includegraphics[width=\textwidth]{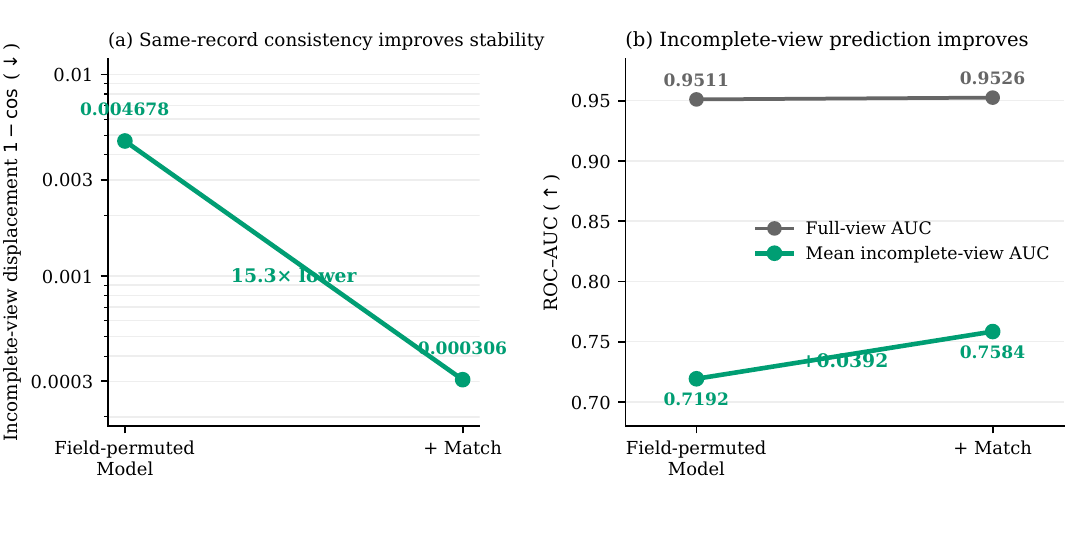}
    \caption{\textbf{Same-record consistency improves incomplete-record stability and prediction when trained with permuted-prefix views.}
    The matched field-permuted \modelstage{} to field-permuted \modelstage{}+\matchstage{} comparison uses
    identical evaluation views and differs only by \matchstage{}. Panel (a) reports incomplete-view
    displacement; lower is better. Panel (b) reports full-view and mean incomplete-view ROC--AUC
    from a logistic probe fitted on full-view embeddings; higher is better. \matchstage{} reduces
    displacement by $15.305\times$ (95\% CI [$15.257\times$, $15.354\times$]) and raises incomplete-view
    AUC by 0.0392 ([0.0233, 0.0555]); the full-view AUC difference is $+0.00145$
    ([$-0.00453$, 0.00763]). MiniBooNE, 3,000 records.}
    \label{fig:match-robustness}
\end{figure}

\subsubsection{Field-dropout protocol and full results}

For each of 3,000 MiniBooNE test records, we first tokenize the raw fixed-order serialization to
512 tokens. The evaluation full view is rebuilt from the maximal prefix of complete key-value units
visible in that encoding; if truncation cuts the next unit, that partial unit is removed. We then
partition only these complete units into two deterministic, complementary strict subsets while
preserving their original order. Thus neither subset can introduce a field absent from its
corresponding full view. The same seed-42 masks are shared by all arms. The
\emph{same-record margin} averages
$\cos(z_A(i),z_B(i))-\cos(z_A(i),z_B(j))$, where the different-record partner $j$ comes from one
fixed seed-7 permutation. Subtracting the cross-record term removes the dominant
record-independent cosine component. \emph{Dropout displacement} is the record mean of
$1-\cos(z_{\mathrm{full}},z_{\mathrm{subset}})$ averaged over both complementary subsets. For task
retention, logistic regression with fixed $C=0.1$ is fit on the first
2,000 full-view embeddings and evaluated on the remaining 1,000 records, once with full
embeddings and once with each subset. ``Half AUC'' is the arithmetic mean of the two subset AUCs;
retention is its ratio to full-view AUC. No alternative masks, pairings, probe splits, or training
seeds are averaged. Probe sensitivity at $C\in\{0.01,0.1,1\}$ is retained in the raw artifact.
The complete protocol, case plan, model hashes, per-record observations, and predictions are stored
in \texttt{learnings/field\_dropout\_corrected\_20260913.json}; its mask-plan SHA-256 is
\texttt{ee700478b20b55d845e61e6c3cfa98272dbc8f70432e4a84cf3c4fc32e5528d4}.
For displacement, paired resamples use all 3,000 records. For AUC, they use the 1,000 held-out records, preserve labels and both views under the same sampled indices, and recompute ROC--AUC in each resample with the fitted probe fixed.

\begin{table}[H]
\caption{Corrected MiniBooNE field-dropout diagnostic. Lower displacement and higher margins indicate
representation stability; full- and mean half-view AUC prevent collapsed representations from
appearing robust. Each half is a strict subset of complete units in its fixed full view. Identical
masks and pairings are used across recipes; the diagnostic contains 3,000 MiniBooNE records. The
displayed 95\% intervals use the paired record-bootstrap protocol above.}
\label{tab:field-dropout}
\centering
\small
\resizebox{\textwidth}{!}{%
\begin{tabular}{lrrcrrcr}
\toprule
Recipe & Same-record margin $\uparrow$ & Displacement $\downarrow$ & Displacement 95\% CI &
Full AUC $\uparrow$ & Half AUC $\uparrow$ & Half AUC 95\% CI & Retention $\uparrow$ \\
\midrule
Frozen decoder & -0.000243 & 0.002419 & [0.002405, 0.002432] & 0.9357 & 0.6937 & [0.6727, 0.7148] & 0.7414 \\
\modelstage{} & -0.002811 & 0.091574 & [0.090951, 0.092195] & 0.9547 & 0.5841 & [0.5583, 0.6094] & 0.6118 \\
\matchstage{} only & 0.000000 & 0.00000306 & [0.00000301, 0.00000310] & 0.8903 & 0.7187 & [0.6999, 0.7372] & 0.8072 \\
\modelstage{} + \matchstage{} & 0.000036 & 0.012001 & [0.011934, 0.012068] & 0.9515 & 0.5331 & [0.5096, 0.5569] & 0.5602 \\
\midrule
Field-permuted \modelstage{} & -0.001106 & 0.004678 & [0.004660, 0.004697] & 0.9511 & 0.7192 & [0.7034, 0.7352] & 0.7562 \\
Field-permuted \modelstage{} + \matchstage{} & -0.000027 & 0.000306 & [0.000305, 0.000307] & 0.9526 & 0.7584 & [0.7420, 0.7754] & 0.7962 \\
\midrule
\modelstage{} $\rightarrow$ \tunestage{} & -0.001142 & 0.013545 & [0.013471, 0.013619] & 0.9435 & 0.5475 & [0.5217, 0.5729] & 0.5803 \\
\modelstage{} + \matchstage{} $\rightarrow$ \tunestage{} & -0.000269 & 0.010406 & [0.010365, 0.010447] & 0.9491 & 0.5377 & [0.5169, 0.5594] & 0.5666 \\
\bottomrule
\end{tabular}}
\end{table}

\subsection{\tunestage{}: target specialization and completion trade-off}

\tunestage{} uses labeled examples to expose the downstream target. It improves the primary
target-prediction metric for every tested Stage-1 parent at 1.5B and remains beneficial for the
field-permuted model-and-match parent through 27B
(Fig.~\ref{fig:matched}(c); Tab.~\ref{tab:model-size-scaling-primary}). Returning to a fixed
field order during \tunestage{} also outperforms retaining field-permuted presentation:
\textsc{Emb-8} rises from 0.741 to 0.753 for the \modelstage{} pipeline and from 0.740 to 0.753
for the model-and-match pipeline (Tab.~\ref{tab:aggregate-ablation-controls}).

This specialization does not improve every decoder capability. Across the four Stage-1 parents,
masked-value completion falls from 0.401, 0.401, 0.448, and 0.445 to 0.145, 0.145, 0.221, and
0.234 after \tunestage{}. The prompt-matched target-emission results in
Appx.~\ref{app:secondary-decoder} confirm that the supervised checkpoints learn to emit their
targets, but no recipe dominates every dataset. Applications requiring both target prediction and
general record completion should therefore retain the Stage-1 checkpoint or use a
retention-aware multi-task objective.

\subsection{Field-order permutation and compute controls}

Field-order permutation is a data intervention rather than an order-invariance test: after
permutation, the fixed token budget exposes different field subsets and local contexts. The
token-budget audit summarized below quantifies this effect, and the selected 1.5B ablations report
the corresponding fixed-order, extended-compute, and field-permuted controls.

Appx.~\ref{app:training-details} summarizes the MiniBooNE token-budget and paired-view audit.
The case-level audit and tokenizer hashes are stored in
\texttt{learnings/view\_truncation\_audit\_20260913.json}. These measurements motivate the term
\emph{permuted-prefix views} and rule out an interpretation as complete, value-preserving
field-order permutations.

\begin{table}[H]
\caption{Selected 1.5B embedding ablations by dataset. \recipedefs}
\centering
\scriptsize
\resizebox{\textwidth}{!}{%
\begin{tabular}{lrrrrrrrrr}
\toprule
Arm & MB & HIGGS & Superc. & USDA & CPU & Naval & Protein & Year & Mean \\
\midrule
Frozen & .956 & .655 & .769 & .910 & .891 & .860 & .311 & .152 & .688 \\
\modelstage{} & .968 & .675 & .808 & .914 & .954 & .920 & .390 & .213 & .730 \\
\modelstage{} + \matchstage{} & .968 & .671 & .813 & .915 & .953 & .923 & .408 & .210 & .733 \\
\matchstage{} only & .941 & .607 & .670 & .869 & .873 & .618 & .242 & .093 & .614 \\
\modelstage{}, 10$\times$ compute & .971 & .691 & .820 & .920 & .930 & .936 & .436 & .245 & .744 \\
Field-permuted \modelstage{} & .967 & .664 & .808 & .942 & .969 & .948 & .396 & .215 & .739 \\
Field-permuted \modelstage{} + \matchstage{} & .969 & .659 & .807 & .937 & .969 & .961 & .420 & .201 & .740 \\
\modelstage{} + order-aware rank & .967 & .685 & .783 & .932 & .952 & .901 & .382 & .195 & .725 \\
\bottomrule
\end{tabular}}
\end{table}

\section{Robustness Across Model Sizes}
\label{app:model-size-results}

\subsection{Model-size robustness from 1.5B to 27B}

The main paper reports compact matched changes across four checkpoint scales. The figure below
visualizes the target-prediction and completion trends, and the subsequent tables provide the
complete values. Because checkpoint architecture, vocabulary, and tokenizer change with scale,
these experiments establish robustness across the evaluated checkpoints rather than a fitted
scaling law.

\begin{figure}[H]
    \centering
    \includegraphics[width=\textwidth]{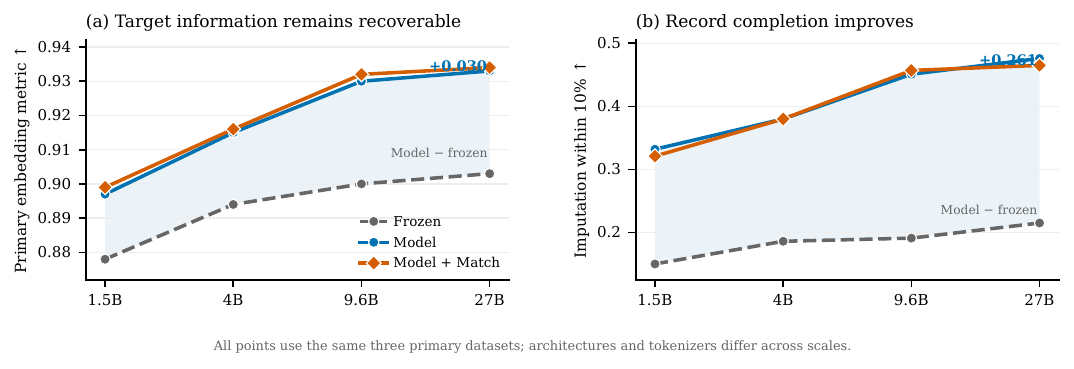}
    \caption{\textbf{Fixed-order \modelstage{} improves target prediction and record completion at
    every tested scale.} Panel (a) reports the unweighted mean of MiniBooNE ROC--AUC,
    Superconductivity $R^2$, and USDA $R^2$; panel (b) reports mean within-10\% masked-value
    imputation on the same datasets. Gray dashed circles denote frozen checkpoints, blue circles
    denote fixed-order \modelstage{}, and orange diamonds add \matchstage{}. Relative to the frozen
    checkpoint, \modelstage{} improves both metrics from 1.5B to 27B; adding \matchstage{} produces
    only small and inconsistent changes. Lines connect separately trained checkpoint families
    whose architectures and tokenizers differ, so this is a one-seed robustness sweep rather than
    a fitted scaling law. \recipedefs}
    \label{fig:scaling}
\end{figure}

Tab.~\ref{tab:model-size-scaling-primary-full} preserves the full per-dataset version of the
compact model-size summary in the main paper. The subsequent tables report the full eight-dataset
ladder available through 9.6B. The 27B experiments were deliberately scoped to the aligned three
primary datasets and are reported in Tabs.~\ref{tab:model-size-scaling-primary-full}
and~\ref{tab:complete-27b-grid}; the two dataset scopes are not mixed in a common mean.

\begin{table*}[t]
\caption{\textbf{Detailed model-size results for selected training recipes on the aligned three
primary datasets.} MiniBooNE reports ROC--AUC, while Superconductivity and USDA report $R^2$;
higher is better throughout, and means are unweighted over the same three datasets. All intended
cells are complete. Because checkpoint architecture and vocabulary also change, this is a
robustness sweep rather than a fitted scaling law. The compact main-paper counterpart is
Tab.~\ref{tab:model-size-scaling-primary}. \recipedefs}
\label{tab:model-size-scaling-primary-full}
\centering
\scriptsize
\setlength{\tabcolsep}{3.2pt}
\renewcommand{\arraystretch}{1.06}
\resizebox{\textwidth}{!}{%
\begin{tabular}{llrrrr}
\toprule
\multirow{2}{*}{Training recipe} & \multirow{2}{*}{Scale}
& \multicolumn{4}{c}{Embedding axis} \\
\cmidrule(lr){3-6}
& & MiniBooNE AUC $\uparrow$ & Supercond. $R^2\uparrow$ & USDA $R^2\uparrow$ & Mean $\uparrow$ \\
\midrule
\multirow{4}{*}{\shortstack[l]{Frozen decoder}}
& 1.5B & .956 & .769 & .910 & .878 \\
& 4B   & .960 & .783 & .940 & .894 \\
& 9.6B & .954 & .807 & .939 & .900 \\
& 27B  & .958 & .804 & .946 & .903 \\
\midrule
\multirow{4}{*}{\shortstack[l]{\modelstage{}}}
& 1.5B & .968 & .808 & .914 & .897 \\
& 4B   & .970 & .830 & .944 & .915 \\
& 9.6B & .974 & .857 & .958 & .930 \\
& 27B  & .974 & .865 & .961 & .933 \\
\midrule
\multirow{4}{*}{\shortstack[l]{\tunestage{} only}}
& 1.5B & .963 & .773 & .939 & .891 \\
& 4B   & .967 & .801 & .964 & .910 \\
& 9.6B & .970 & .831 & .978 & .926 \\
& 27B  & .966 & .836 & .978 & .927 \\
\midrule
\multirow{4}{*}{\shortstack[l]{\modelstage{} $\rightarrow$ \tunestage{}}}
& 1.5B & .969 & .804 & .943 & .905 \\
& 4B   & .968 & .821 & .965 & .918 \\
& 9.6B & .967 & .858 & .972 & .932 \\
& 27B  & .970 & .849 & .974 & .931 \\
\midrule
\multirow{4}{*}{\shortstack[l]{Field-permuted \modelstage{} + \matchstage{}}}
& 1.5B & .969 & .807 & .937 & .904 \\
& 4B   & .970 & .841 & .954 & .922 \\
& 9.6B & .971 & .864 & .966 & .934 \\
& 27B  & .970 & .868 & .970 & .936 \\
\midrule
\multirow{4}{*}{\shortstack[l]{Field-permuted \modelstage{} + \matchstage{}\\$\rightarrow$ \tunestage{}}}
& 1.5B & .968 & .806 & .967 & .914 \\
& 4B   & .971 & .847 & .973 & .930 \\
& 9.6B & .973 & .851 & .980 & .935 \\
& 27B  & .973 & .873 & .979 & .942 \\
\bottomrule
\end{tabular}}
\end{table*}

\begin{table}[H]
\caption{Full eight-dataset embedding results available through 9.6B: AUC for MiniBooNE/HIGGS,
$R^2$ otherwise. \recipedefs}
\label{tab:model-size-embedding-full}
\centering
\scriptsize
\resizebox{\textwidth}{!}{%
\begin{tabular}{llrrrrrrrrr}
\toprule
Model & Post-training state & MB & HIGGS & Superc. & USDA & CPU & Naval & Protein & Year & Mean \\
\midrule
1.5B & Frozen & .956 & .655 & .769 & .910 & .891 & .860 & .311 & .152 & .688 \\
1.5B & \modelstage{} & .968 & .675 & .808 & .914 & .954 & .920 & .390 & .213 & .730 \\
4B & Frozen & .960 & .642 & .783 & .940 & .926 & .893 & .332 & .160 & .705 \\
4B & \modelstage{} & .970 & .678 & .830 & .944 & .961 & .951 & .433 & .227 & .749 \\
9.6B & Frozen & .954 & .644 & .807 & .939 & .928 & .897 & .349 & .149 & .708 \\
9.6B & \modelstage{} & .974 & .739 & .857 & .958 & .970 & .986 & .506 & .240 & .779 \\
\bottomrule
\end{tabular}}
\end{table}

\begin{table}[H]
\caption{Full eight-dataset masked-value-imputation results available through 9.6B. \recipedefs}
\centering
\scriptsize
\resizebox{\textwidth}{!}{%
\begin{tabular}{llrrrrrrrrr}
\toprule
Model & Post-training state & MB & HIGGS & Superc. & USDA & CPU & Naval & Protein & Year & Mean \\
\midrule
Median floor & -- & .242 & .232 & .276 & .134 & .242 & .238 & .282 & .048 & .212 \\
1.5B & Frozen & .004 & .126 & .250 & .196 & .158 & .086 & .000 & .028 & .106 \\
1.5B & \modelstage{} & .214 & .288 & .494 & .288 & .366 & .914 & .576 & .070 & .401 \\
4B & Frozen & .006 & .130 & .292 & .260 & .194 & .144 & .000 & .024 & .131 \\
4B & \modelstage{} & .226 & .296 & .572 & .342 & .388 & .914 & .618 & .076 & .429 \\
9.6B & Frozen & .006 & .172 & .280 & .286 & .174 & .122 & .000 & .034 & .134 \\
9.6B & \modelstage{} & .316 & .326 & .662 & .376 & .394 & .912 & .646 & .070 & .463 \\
\bottomrule
\end{tabular}}
\end{table}

\subsection{Masked-value completion across model sizes}

The eight-dataset table above gives the broad frozen-versus-\modelstage{} comparison through
9.6B. The following aligned three-dataset table includes additional training recipes and the
completed 27B checkpoints; the overlapping cells are retained to keep each scope self-contained.
Masked-value imputation asks the decoder to regenerate a field value that has been blanked from the
serialized record; a prediction counts as correct when it lands within 10\% of the true value. These
numbers were previously carried alongside the embedding axis in
Tab.~\ref{tab:model-size-scaling-primary}; they are reported here instead to keep the main table
focused on downstream prediction. Higher is better and the mean is unweighted over the same three
primary datasets at every scale. \recipedefs

\begin{table*}[t]
\caption{\textbf{Masked-value imputation across model sizes for selected training recipes.}
Values are within-10\% accuracy on the aligned three primary datasets; higher is better, and the
mean is unweighted. All intended cells are complete. }
\label{tab:model-size-imputation-full}
\centering
\scriptsize
\setlength{\tabcolsep}{4pt}
\begin{tabular}{llrrrr}
\toprule
Training recipe & Scale & MiniBooNE $\uparrow$ & Supercond. $\uparrow$ & USDA $\uparrow$ & Mean $\uparrow$ \\
\midrule
Frozen decoder & 1.5B & .004 & .250 & .196 & .150 \\
                                & 4B   & .006 & .292 & .260 & .186 \\
                                & 9.6B & .006 & .280 & .286 & .191 \\
                                & 27B  & .008 & .350 & .288 & .215 \\
\midrule
\modelstage{}            & 1.5B & .214 & .494 & .288 & .332 \\
                                & 4B   & .226 & .572 & .342 & .380 \\
                                & 9.6B & .316 & .662 & .376 & .451 \\
                                & 27B  & .316 & .680 & .432 & .476 \\
\midrule
\tunestage{} only    & 1.5B & .016 & .132 & .090 & .079 \\
                                & 4B   & .014 & .174 & .108 & .099 \\
                                & 9.6B & .010 & .096 & .048 & .051 \\
                                & 27B  & .014 & .082 & .100 & .065 \\
\midrule
\modelstage{} $\rightarrow$ \tunestage{} & 1.5B & .022 & .168 & .090 & .093 \\
                                & 4B   & .014 & .168 & .146 & .109 \\
                                & 9.6B & .016 & .120 & .082 & .073 \\
                                & 27B  & .040 & .072 & .170 & .094 \\
\midrule
Field-permuted \modelstage{} + \matchstage{} & 1.5B & .034 & .138 & .060 & .077 \\
\quad $\rightarrow$ \tunestage{}   & 4B   & .032 & .202 & .022 & .085 \\
                                & 9.6B & .034 & .118 & .042 & .065 \\
                                & 27B  & .218 & .112 & .020 & .117 \\
\bottomrule
\end{tabular}
\end{table*}

Only the frozen decoder and \modelstage{} improve monotonically with scale here; every
\tunestage{} recipe remains well below \modelstage{} alone, which is the expected cost of
specializing the decoder to emit a single target rather than to model the whole record.

\subsection{Complete 27B recipe grid}

The complete grid separates the effects of \modelstage{}, \matchstage{}, \tunestage{},
field-order permutation, and additional optimization on the default serialization at the largest
evaluated scale.

\begin{table}[H]
\caption{Complete 27B grid on the aligned three-primary-dataset scope. The embedding column is the
mean primary downstream metric; imputation is within-10\% accuracy. Higher is better. }
\label{tab:complete-27b-grid}
\centering
\scriptsize
\setlength{\tabcolsep}{3.5pt}
\resizebox{\columnwidth}{!}{%
\begin{tabular}{lllrr}
\toprule
Recipe & \modelstage{} view & \matchstage{} / \tunestage{} & Embedding $\uparrow$ & Imputation $\uparrow$ \\
\midrule
Frozen decoder & -- & -- & .903 & .215 \\
\modelstage{} & default view & -- & .933 & .476 \\
\modelstage{} + \matchstage{} & default view & \matchstage{} & .934 & .465 \\
\tunestage{} only & -- & \tunestage{} only & .927 & .065 \\
Extended-compute \modelstage{} & default view, 10$\times$ & -- & .920 & .479 \\
Field-permuted \modelstage{} & permuted-prefix views & -- & .936 & \textbf{.613} \\
Field-permuted \modelstage{} + \matchstage{} & permuted-prefix views & \matchstage{} & .936 & .604 \\
\modelstage{} $\rightarrow$ \tunestage{} & default view & \tunestage{} & .931 & .094 \\
\modelstage{} + \matchstage{} $\rightarrow$ \tunestage{} & default view & \matchstage{}$\rightarrow$\tunestage{} & .928 & .075 \\
Field-permuted \modelstage{} + \matchstage{} $\rightarrow$ \tunestage{} & permuted-prefix views & \matchstage{}$\rightarrow$\tunestage{} & \textbf{.942} & .117 \\
\bottomrule
\end{tabular}}
\end{table}
Across all four tested checkpoints, \modelstage{} obtains higher embedding and masked-value-imputation aggregates than the corresponding frozen checkpoint. The complete recipe obtains higher target-prediction scores than \tunestage{} alone at all four tested checkpoints and has the highest reported 27B embedding score (0.942).
At equal update budgets, field-permuted \modelstage{} also exceeds extended-compute \modelstage{} at 27B by 0.016 on embedding and 0.134 on imputation. In contrast, adding \matchstage{} to \modelstage{}, field-permuted \modelstage{}, or
\modelstage{}$\rightarrow$\tunestage{} remains aggregate-inert or negative through 27B, and \matchstage{} does not improve imputation over its Stage-1 parent. Thus, \matchstage{} contracts paired views but is not a source of scale-growing aggregate accuracy; whether that contraction is useful remains view-dependent.



\section{Additional Direct-Decoder Evaluations}
\label{app:secondary-decoder}

\subsection{Evaluation scope and prompt compatibility}
The currently reported held-out record perplexity, masked-value imputation, and field-lookup artifacts
all use the original decoder harness's code-level 320-token input cap. Target emission below is a
separate diagnostic invocation: it uses the \tunestage{} training template
(\texttt{--g1\_prompt train}) and a 496-token input cap (\texttt{--g1\_max\_input 496}). It is
evaluated only on the four datasets for which the complete prompt is valid. The direct decoder
aggregate in the main paper is eight-dataset masked-value imputation; target emission is reported
separately on this prompt-valid subset.

\subsection{Prompt-matched target emission}
Target emission asks the decoder to produce the dataset target from the available fields. This is a
secondary, prompt-conditioned diagnostic rather than the eight-dataset masked-value-imputation axis.
Tab.~\ref{tab:target-emission-clean} reports only the four datasets whose complete Stage-2 prompt,
including the \texttt{Question: ... Answer:} suffix, fits within the 496-token \tunestage{} budget. HIGGS uses
classification accuracy (higher is better); the three regression datasets use median absolute log
ratio, $\operatorname{median}|\log(\hat y/y)|$ (lower is better). Under the prompt used during \tunestage{},
all four \tunestage{} recipes beat the task-specific floor on all four datasets, but no recipe
dominates: \modelstage{}+\matchstage{}$\rightarrow$\tunestage{} is best on HIGGS, CPU, and
Protein, while the field-permuted complete recipe is best on Naval.

\begin{table}[H]
\caption{Corrected 1.5B target-emission results on the four prompt-clean datasets. The prompt matches
\tunestage{} training and uses a 496-token budget. Bold marks the best \tunestage{} recipe per dataset. ``n/p'' denotes
no parseable numerical output, with the percentage of 500 evaluated records in parentheses. \recipedefs}
\label{tab:target-emission-clean}
\centering
\small
\resizebox{\textwidth}{!}{%
\begin{tabular}{lrrrr}
\toprule
Recipe & HIGGS Acc. $\uparrow$ & CPU $\downarrow$ & Naval $\downarrow$ & Protein $\downarrow$ \\
\midrule
Task-specific floor & 0.518 & 0.086 & 0.013 & 0.755 \\
Frozen decoder & 0.091 & n/p (100\%) & n/p (89\%) & n/p (100\%) \\
\modelstage{} & 0.014 & n/p (100\%) & n/p (100\%) & n/p (100\%) \\
\midrule
\tunestage{} only & 0.720 & 0.013 & 0.002 & 0.310 \\
\modelstage{} $\rightarrow$ \tunestage{} & 0.714 & 0.022 & 0.002 & 0.301 \\
\modelstage{} + \matchstage{} $\rightarrow$ \tunestage{} & \textbf{0.722} & \textbf{0.012} & 0.002 & \textbf{0.294} \\
Field-permuted \modelstage{} + \matchstage{} $\rightarrow$ \tunestage{} & 0.708 & 0.015 & \textbf{0.001} & 0.341 \\
\bottomrule
\end{tabular}}
\end{table}

MiniBooNE (718 tokens), USDA
FoodData (597), YearPredictionMSD (977), and Superconductivity (1110) exceed the 496-token Stage-2
budget; right truncation deleted the question-and-answer suffix, so evaluating those checkpoints on
that suffix would measure a train/test prompt mismatch. We exclude all four rather than treating
their outputs as model failures. The remaining four datasets have maximum serializations of at most
362 tokens. The result is also prompt-sensitive. With the original 320-token instruction-style
harness, \tunestage{} only instead wins HIGGS, while the field-permuted complete recipe wins CPU,
Naval, and Protein. All \tunestage{} recipes still beat the corresponding floors on HIGGS, Naval,
and Protein, but \tunestage{} only and \modelstage{}$\rightarrow$\tunestage{} miss the CPU floor;
the changed outcomes prevent a general recipe ranking. The frozen decoder and \modelstage{} were
never trained on the terse \tunestage{} template and frequently emit prose rather than a parseable
number, as the coverage entries in Tab.~\ref{tab:target-emission-clean} show. 

\end{document}